\documentclass{article}

\usepackage{microtype}
\usepackage{graphicx}
\usepackage{booktabs} %

\usepackage{hyperref}

\usepackage[accepted]{mlsys2023}

\mlsystitlerunning{Validating Large Language Models with ReLM}

\usepackage{hyperref}
\usepackage{xfrac}
\usepackage{subcaption}
\usepackage{graphicx}
\usepackage{booktabs} %
\usepackage{xcolor}
\usepackage{wrapfig}
\usepackage{adjustbox}
\usepackage{tabularx}
\usepackage{makecell,rotating}
\usepackage{enumitem}
\usepackage{amsthm}
\usepackage{amsmath}
\usepackage{xspace}
\usepackage{soul}
\usepackage[utf8]{inputenc}
\usepackage{blindtext}
\usepackage[htt]{hyphenat}
\usepackage{tikz} %
\graphicspath{{Figures/}}
\usepackage{xurl}
\usepackage{bm}
\DeclareFixedFont{\ttb}{T1}{txtt}{bx}{n}{12} %
\DeclareFixedFont{\ttm}{T1}{txtt}{m}{n}{12}  %

\usepackage{color}
\definecolor{deepblue}{rgb}{0,0,0.5}
\definecolor{deepred}{rgb}{0.6,0,0}
\definecolor{deepgreen}{rgb}{0,0.5,0}
\definecolor{codegreen}{rgb}{0,0.6,0}
\definecolor{codegray}{rgb}{0.5,0.5,0.5}
\definecolor{codepurple}{rgb}{0.58,0,0.82}
\definecolor{backcolour}{rgb}{0.95,0.95,0.92}
\definecolor{codeorange}{rgb}{1.0,0.64,0.0}

\usepackage{listings}

\newcommand\digitstyle{\color{codeorange}}
\makeatletter
\newcommand{\ProcessDigit}[1]
{%
  \ifnum\lst@mode=\lst@Pmode\relax%
   {\digitstyle #1}%
  \else
    #1%
  \fi
}

\newcommand\pythonstyle{\makeatother\lstset{
language=Python,
basicstyle=\footnotesize\ttm,
deletekeywords={map},
otherkeywords={self},             %
emph={MyClass,__init__},          %
emphstyle=\color{deepred},    %
stringstyle=\color{deepgreen},
frame=tb,                         %
showstringspaces=false,            %
    commentstyle=\color{codegray},
    keywordstyle=\color{magenta},
    numberstyle=\tiny\color{codepurple},
    stringstyle=\color{codegreen},
    basicstyle=\ttfamily\footnotesize,
    breakatwhitespace=true,
    breaklines=true,
    captionpos=b,
    keepspaces=true,
    numbers=left,
    numbersep=5pt,
    showspaces=false,
    showtabs=false,
    tabsize=2,
    keywordstyle=[2]{\color{codeorange}},
    morekeywords=[2]{None},
  literate=
    {0}{{{\ProcessDigit{0}}}}1
    {1}{{{\ProcessDigit{1}}}}1
    {2}{{{\ProcessDigit{2}}}}1
    {3}{{{\ProcessDigit{3}}}}1
    {4}{{{\ProcessDigit{4}}}}1
    {5}{{{\ProcessDigit{5}}}}1
    {6}{{{\ProcessDigit{6}}}}1
    {7}{{{\ProcessDigit{7}}}}1
    {8}{{{\ProcessDigit{8}}}}1
    {9}{{{\ProcessDigit{9}}}}1
    {<=}{{\(\leq\)}}1,
    morestring=[b]",
    morestring=[b]',
    morecomment=[l]//,
}}

\lstnewenvironment{python}[1][]
{
\pythonstyle
\lstset{#1}
}
{}

\newcommand\pythonexternal[2][]{{
\pythonstyle
\lstinputlisting[#1,mathescape=true]{#2}}}

\newcommand\pythoninline[1]{{\pythonstyle\lstinline!#1!}} %

\newcommand{\relm}{\texttt{ReLM}\xspace}
\newcommand{\warningOffensive}{{\color{orange}WARNING: This section contains
examples which are offensive in nature.}}
\newcounter{inlineenum}
\renewcommand{\theinlineenum}{\arabic{inlineenum}}
\newenvironment{inlineenum}
  {\unskip\ignorespaces\setcounter{inlineenum}{0}%
   \renewcommand{\item}{\refstepcounter{inlineenum}{\theinlineenum)~}}}
  {\ignorespacesafterend}
\newcommand*\circled[2][1.6]{\tikz[baseline=(char.base)]{
  \node[shape=circle, draw, inner sep=1pt,
      minimum height={\f@size*#1},] (char) {\vphantom{WAH1g}#2};}}

\newcounter{obscount}
\newcommand{\observation}{ %
\addtocounter{obscount}{1} %
\textbf{Observation \arabic{obscount}: }}
\definecolor{bubblegum}{rgb}{0.99, 0.76, 0.8}
\definecolor{cherryblossompink}{rgb}{1.0, 0.72, 0.77}
\definecolor{flamingopink}{rgb}{0.99, 0.56, 0.67}

\newcommand\mkprogress[1]{}
\newcommand\code[1]{{\color{deepgreen}\texttt{#1}}}
\newcommand\nofontcode[1]{{\color{deepgreen}#1}}
\newcommand\query[1]{{\color{deepgreen}\texttt{#1}}}
\newcommand\token[1]{{\texttt{#1}}}
\newcommand\relmstring[1]{{\color{deepgreen}#1}}
\newcommand\EOSToken{\circled[0.25]{{\textsc{\tiny{Eos}}}}}

\def\vx{{\bm{x}}}
 
\usepackage[firstpage]{draftwatermark}
\SetWatermarkText{
 \hspace*{3.5in}
 \raisebox{10.0in}{
  \includegraphics[height=0.9in]{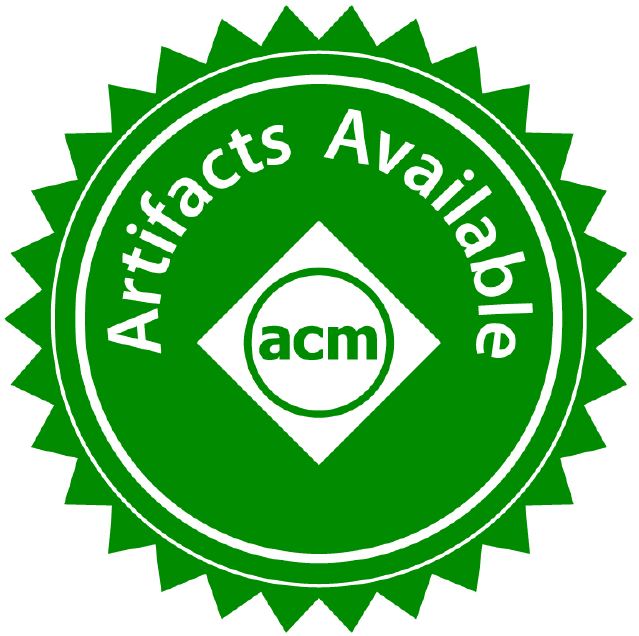}
  \includegraphics[height=0.9in]{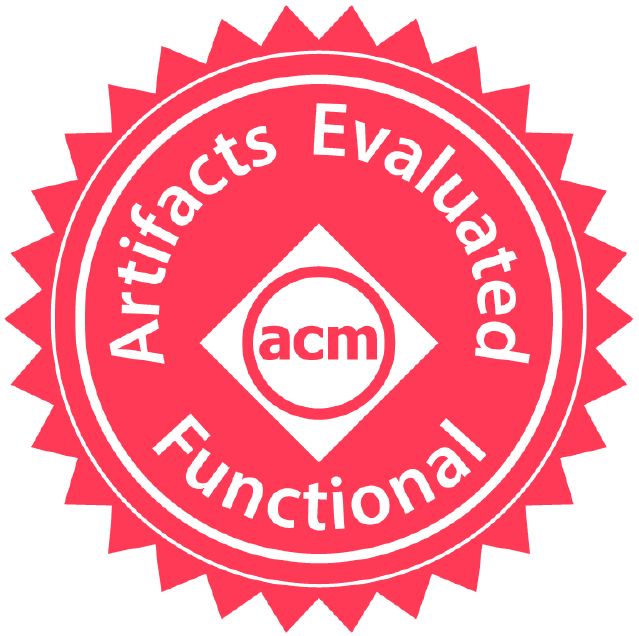}
 }
}
\SetWatermarkAngle{0}
 
\begin{document}

\twocolumn[
\mlsystitle{Validating Large Language Models with ReLM}

\mlsyssetsymbol{equal}{*}

\begin{mlsysauthorlist}
\mlsysauthor{Michael Kuchnik}{cmu}
\mlsysauthor{Virginia Smith}{cmu}
\mlsysauthor{George Amvrosiadis}{cmu}
\end{mlsysauthorlist}

\mlsysaffiliation{cmu}{Carnegie Mellon University}

\mlsyscorrespondingauthor{Michael Kuchnik}{mkuchnik@cmu.edu}

\mlsyskeywords{Machine Learning, MLSys, Regular Expression, Validation, Large Language Model}

\vskip 0.3in

\begin{abstract}
Although large language models (LLMs) have been touted for their ability to
generate natural-sounding text, there are growing concerns around possible
negative effects of LLMs such as data memorization, bias, and inappropriate
language.
Unfortunately, the complexity and generation capacities of LLMs make validating
(and correcting)
such concerns difficult.
In this work, we introduce \relm, a system for validating and querying LLMs
using standard regular expressions.
\relm{} formalizes and enables a broad range of language model evaluations,
reducing complex evaluation rules to simple regular expression queries.
Our results exploring queries surrounding memorization, gender bias, toxicity, and
language understanding
show that \relm{} achieves up to $15\times$ higher
system efficiency, $2.5\times$ data efficiency, and increased
statistical and prompt-tuning coverage compared to state-of-the-art ad-hoc
queries.
\relm{} offers a competitive and general baseline for the increasingly important problem of LLM validation.

 \end{abstract}
]

\printAffiliationsAndNotice{}  %

\section{Introduction}%
\label{sec:introduction}%
Large language models (LLMs), such as GPT-3~\cite{GPT3} and PaLM~\cite{PALM},
are a popular tool for many natural language processing tasks.
While it is well understood that these models are expensive to train and deploy,
there are growing concerns around even the seemingly simple problem of
\textit{validating} LLM
behavior~\cite{what_will_it_take_fix_nlp,srivastava2022beyond}.
Validation is particularly important in that LLMs may have unintended effects~\cite{bommasani2021opportunities},
such as returning memorized training data~\cite{carlini2022quantifying},
encoding bias in results~\cite{bender2021dangers},
and generating inappropriate
content~\cite{gehman-etal-2020-realtoxicityprompts,ousidhoum-etal-2021-probing,GPT3}.
While there have been extensive efforts to perform such testing on
LLMs, the tests are written in an ad-hoc manner, where test
maintainers explicitly code out the test's logic using LLM-specific
utilities~\cite{kiela-etal-2021-dynabench,srivastava2022beyond,eval-harness}.
Test users then add tasks by extending the existing code or supplying
template parameters for that code (e.g., via millions of lines of JSON).
In these approaches, it is up to the user to convert the expected LLM behavior into a
sequence of test vectors that can be executed, making it difficult to
maintain, modify, and extend test functionality.

\begin{figure*}%
  \begin{subfigure}[b]{0.33\linewidth}
    \centering
    \includegraphics[width=\linewidth]{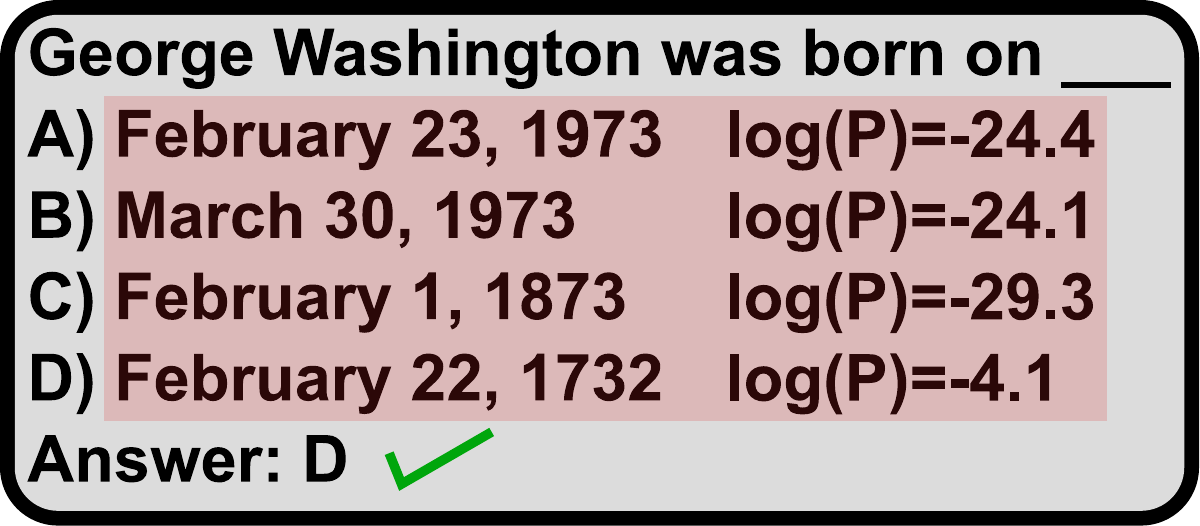}%
    \caption{Multiple Choice}%
    \label{fig:mc_response}%
  \end{subfigure}%
  \hfill%
  \begin{subfigure}[b]{0.33\linewidth}
    \centering
    \includegraphics[width=\linewidth]{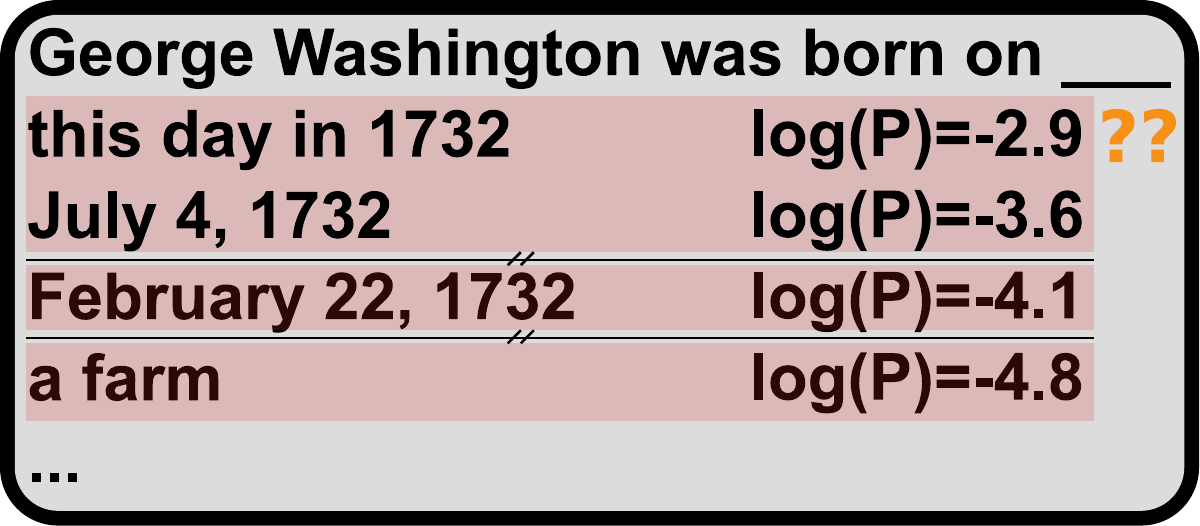}%
    \caption{Free Response}%
    \label{fig:free_response}%
  \end{subfigure}%
  \hfill%
  \begin{subfigure}[b]{0.33\linewidth}
    \centering
    \includegraphics[width=\linewidth]{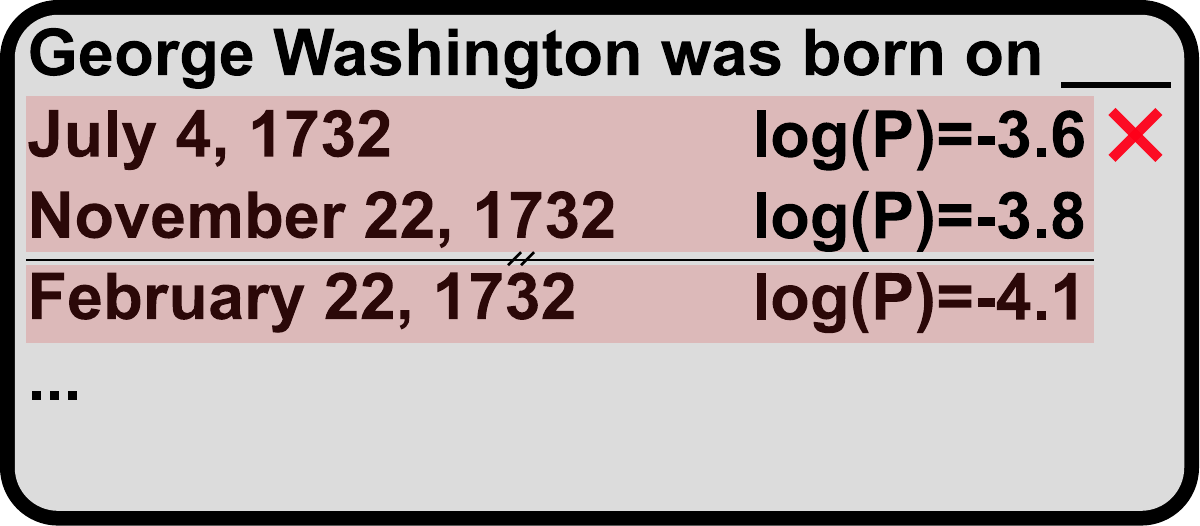}%
    \caption{Any Date}%
    \label{fig:any_date}%
  \end{subfigure}%
  \caption{Testing an LLM's knowledge of George Washington's birth date (LLM
  predictions highlighted in {\color{flamingopink}pink}).
    (\ref{fig:mc_response}) Using 4 out of all possible dates and ranking them.
    An LLM classifying on year alone is sufficient to guess the answer correctly,
    limiting test resolution.
    Note that the answers can alternatively be encoded in the prompt, with
    predictions over the answer's letter.
    (\ref{fig:free_response}) Allowing the LLM to output any completion,
    resulting in unexpected responses.
    (\ref{fig:any_date}) A structured query over all dates of the form
    \code{\textbf{<Month> <Day>, <Year>}}, obtaining the specificity of
    \ref{fig:mc_response} with the
    generality of \ref{fig:free_response}.
    Our approach finds that, among a search space of all dates,
    GPT-2XL's highest ranked prediction is incorrect, though the correct prediction
    is in the top 10.
    The same approach shows that the small variant of GPT-2 cannot discern the
    date even in \ref{fig:mc_response}.
  }%
  \label{fig:questions_overview}%
  \vspace{-10pt}%
\end{figure*}

As an example, consider testing if an LLM knows the birth date of George
Washington via a fill-in-the-blank query:
\texttt{George Washington was born on \rule{1cm}{0.15mm}}, as shown in
Figure~\ref{fig:questions_overview}.
Today's standard solution is to prompt the model with a
``multiple-choice'' assessment using a few dates~\cite{srivastava2022beyond}
(Figure~\ref{fig:mc_response}).
However, such ``closed-choice'' assessments have shortcomings:
with 12 months in a year, 31 days in a month, and thousands of years to
consider, there are %
millions of candidates, and thus, the selection of dates can bias the
evaluation.
For example, the selected answer will always change if a more probable candidate was
introduced, making the test prone to false positives.
The alternative, sampling open-domain ``free-response'' strings
(Figure~\ref{fig:free_response}),
is more challenging to test, as it requires grading arbitrary strings relative
to
all representations of the correct
date.
While the former case may bias performance, the latter seems impossible to
control: every possible response, including \texttt{this day in 1732}
or \texttt{a farm},
must be considered and graded, which requires carefully tuning the evaluation to
avoid false positives and false
negatives~\cite{Prager2006OpenDomainQ,roberts-etal-2020-much}.
The inability to control the response of the LLM is in itself a testing bottleneck,
motivating structured queries (Figure~\ref{fig:any_date}), which are
\textit{guaranteed}
to be drawn from the full set of expected strings (e.g., those of the pattern
\code{\textbf{<Month> <Day>, <Year>}}) without unexpected responses.

In this work, we introduce the first queryable test interface for LLMs.
Our system, \relm, is a \underline{R}egular \underline{E}xpression engine
for \underline{L}anguage \underline{M}odels and enables
executing commonly-occurring \textit{patterns}---sets of strings---with \textit{standard regular expressions}.
\relm{} is the first system expressing a query as the complete \textit{set} of
test patterns, empowering practitioners to directly measure LLM behavior over
sets too large to enumerate.
The key to \relm{}'s success is its ability to compactly represent the solution
space via a graph representation, which is derived from regular expressions,
compiled to an LLM-specific representation, and finally executed.
As a result, users do not have to understand implementation details of the LLM itself---tests
have the same effect as if all string possibilities were materialized.
In addition to introducing \relm{}, we
demonstrate how various LLM evaluation tasks can be mapped onto \relm's string
patterns.

For \relm{} users, programming a validation task consists of two components:
\begin{inlineenum}
\item formally specifying a set of strings of interest via a regular expression and
\item instructing the engine on how to enumerate and score the strings\@.
\end{inlineenum}
For example, memorization tests can be expressed as finding a sample of training data in the LLM,
bias can be expressed as the co-occurrence of data in sampled LLM sentences,
toxicity can be expressed as finding insults within LLM-generated sentences,
and
knowledge can be expressed by assigning higher likelihood to the correct answer.
However, unlike enumerating the sequences in a pattern,
\relm's queries are \textit{succinctly} defined with a graph representation,
allowing \relm{} to scale to queries with
billions of strings in a few lines of code.
Using the examples of URL memorization, gender bias, toxicity, and
language understanding tasks with
GPT-2
models, we demonstrate that \relm{} is both efficient
and expressive in executing common queries---dramatically lowering the bar to validate LLMs.
Our contributions are:

\vspace{3pt}
\begin{enumerate}[nosep,wide=0pt,label={\textbf{(\arabic*)}}]%
\setlength{\itemsep}{3pt}
    \item A formalization of regular expressions on LLM predictions.
      Unlike multiple choice questions, which are few and enumerable, regular
      expressions can model sets of infinite size.
      Unlike free response questions, which may result in spurious answers,
      \relm's results are always well-defined.
    \item An identification and construction of two commonly used classes of LLM
      inference queries, \textit{conditional} and \textit{unconditional}
      generation.
      For unconditional generation, we find that a fixed query string can be
      represented by many token sequences, motivating a compressed
      representation.
      To the best of our knowledge, we are the first to support
      these alternative encodings via automata.
    \item The design and implementation of a regular expression to inference
engine, which efficiently maps regular expressions to finite automata.
    We implement both shortest path and randomized graph traversals that
    yield outputs in seconds at competitive GPU utilizations.
    \item An evaluation of memorization, gender bias, toxicity, and language understanding
      tasks using GPT-2
      models, where we demonstrate
      the utility of \relm{} within the context of LLM validation.
    \relm{} achieves a $15\times$ speedup or $2.5\times$ higher data
    efficiency over traditional
    approaches in memorization and toxicity finding, respectively.
    As a diagnostic tool, \relm{}
    exposes testing over character- and token-level transformations,
    measuring the robustness of bias to input representations.
    As a tuning tool, \relm{}
    effortlessly supports common prompt tuning transformations necessary for state-of-the-art
    zero-shot performance, enabling practitioners to quickly iterate on their
    prompt design.
\end{enumerate}

\section{Background and Related Work}%
\label{sec:background}%
This work, being a regular expression engine for LLMs, primarily spans
classical formal language theory as well as modern LLM architectures.
We motivate the problem of LLM validation  in
Section~\ref{sec:nlp_background} and then discuss how prompts and tests are structured in
Section~\ref{sec:prompts_background}.
In Section~\ref{sec:formal_languages_background},
we re-cast the specification of LLM prompts and LLM outputs
using formal languages.
Finally, we focus on the semantics of testing
for particular behavior in autoregressive models in Section~\ref{sec:lm_validation_background}.

\subsection{The Shifting Landscape of LLM Validation}%
\label{sec:nlp_background}%
The Transformer architecture~\cite{vaswani2017attention} resulted
in the backbone of many LLMs.
Autoregressive models, such as the GPT family~\cite{GPT1,GPT2,GPT3}, were able
to adapt to downstream tasks by representing the task specification itself
inside the input (i.e., in the prompt)~\cite{rocktaschel2015reasoning}.
Masked models, of the BERT~\cite{BERT} family, instead filled the role of
transfer via fine-tuning.
Subsequent work unified the interface of transfer-focused models such that all inputs and
outputs are strings~\cite{JMLR:T5}.
Recently, LLMs have exceeded human capabilities on many
benchmarks~\cite{superglue}, raising concerns that standard benchmarks are no longer sufficient
for tracking progress in the field~\cite{what_will_it_take_fix_nlp,srivastava2022beyond}.
These trends point to a need for more rigorous and holistic validation
efforts~\cite{helm},
where an LLM's performance is measured over tasks spanning:
\begin{inlineenum}
\item strings in both input and output space,
\item enough difficult content to generate a ``grading curve'', and
\item a variety of model behavior such that performance is broken
  down by area (e.g., memorization, bias, toxicity, knowledge).
\end{inlineenum}
There are currently primarily two ways to grade LLMs in a black-box fashion: \textit{multiple choice}
and \textit{free response} questions~\cite{srivastava2022beyond}, which we discuss formally below (\S\ref{sec:lm_validation_background}).
Multiple choice questions present typically 2--10 completions to the LLM, which are scored,
and the most likely response is used as the LLM's answer.
Free response questions allow the LLM to generate any completion.
In both cases, the LLM's answer is checked (usually verbatim) against a reference solution to
assign a score.
LLM validation is thus analogous to software-engineering's unit-tests over
strings in the LLM's output space (e.g., over $10^{4000}$ for GPT-2).

\subsection{Specifying the Input/Output of LLMs}%
\label{sec:prompts_background}%
In LLMs, both the inputs and outputs of the model are often strings.
The use of strings as a data representation makes it possible to form
predictions over mostly arbitrary objects, by simply converting their
representation into a string form.
The act of forming a string input is called \textit{prompt engineering} and is an active research
area~\cite{gao-etal-2021-making,liu2021gpt,schick-schutze-2021-exploiting,jiang-etal-2020-know,promptprogramming}.
The act of testing a string output has many names, but most forms of testing can
be viewed as generalizations of fill-in-the-blank tests.
These tests, introduced as \textit{cloze tests} in human
psychology~\cite{Taylor1953ClozePA}, were used to measure human aptitude in understanding
and reasoning about context, and can similarly be used for LLMs.
The use of a cloze is a training primitive
for masked LLMs, where randomly selected words are masked out and
\underline{predicted}, as well as autoregressive LLMs, which have a
\textit{causal} constraint on the mask~\cite{JMLR:T5}.
The design of tests is itself an active research area; one major focus~\cite{kiela-etal-2021-dynabench} has been
to disregard randomly sampled natural data and focus mostly on adversarial
inputs---inputs that have been analytically or empirically found to fool
certain models.
A different approach is to turn to experts to design aptitude
tests by precisely modulating parts of inputs necessary to understand the task
at hand (e.g., the digits of a number in addition
tasks)~\cite{sugawara2020assessing,dunietz-etal-2020-test}.
Others advocate for large (or difficult), precisely constructed datasets that
also test bias~\cite{what_will_it_take_fix_nlp}.

\subsection{Expressing Input/Output via Formal Languages}%
\label{sec:formal_languages_background}%
For many LLM tasks, there is precise definition of input/output relations i.e.,
a pattern that matches on a set of input/output strings, which are studied under formal languages.
An alphabet, $\Sigma$, is a finite set of symbols.
Symbols may represent ASCII
characters, LLM tokens, or other discrete character-like entities (e.g., emojis
or states).
Strings are lists of symbols, and a language $L$ over an alphabet $\Sigma$ represents a set of strings out of all
possible strings $\Sigma^*$ in $\Sigma$.
A fundamental pattern for defining languages is the family of \textit{regular
expressions}~\cite{automatalanguages}, which extend string literals (a concatenation of symbols) to more
operations, namely disjunction (e.g., $a|b$) and zero or more repetitions
(e.g., $a^*$).
A regular expression is equivalent to a \textit{finite-state automaton}, a directed graph
representing valid transitions from a start state to end states.
A finite-state automaton is defined by $Q$, the set of states, $\Sigma$, the set
of input symbols, $\delta$, the transition function over $Q \times \Sigma \to Q$,
$q_0 \in Q$, the  initial state, and $F \subseteq Q$, the set of final states.
Conversion from regular expressions to automata is covered by textbooks~\cite{automatalanguages}.
\textit{Transducers} are extensions of automata that have output
symbols and weights at each edge, mapping from one language to another.
Algebraic operations, like difference, intersection, and composition, can be used
to transform languages
abstractly~\cite{mohri-1997-finite,SpeechRecognitionComposition}.
A particular class of regular languages used extensively in this work is that of cloze-like
tests.
If a prompt or premise consists of some string, $\alpha$, followed by a pattern
or mask,
$\beta$, then the test operates over the language defined by their
concatenation $L=\alpha\beta$.

\subsection{Testing LLMs with Formal Languages}%
\label{sec:lm_validation_background}%
A language model assigns probabilities over vast sets of strings---some are even designed to
be Unicode-complete~\cite{GPT2}.
As tests are rarely defined over raw probabilities, there must be a \textit{decision
rule} to convert probabilities into a binary choice over strings.
\textit{Autoregressive} LLMs, which we focus on, form a total
probability by iteratively predicting the next token of a string, from left
to right: $p(x_1,x_2,\dots,x_n) = \prod_{i=1}^n p(x_i|x_1,x_2,\dots,x_{i-1})$, where $x_i$ is
a token representing characters, subwords, or whole words~\cite{BPE,GPT2} in a
sequence $\vx=x_1,x_2,\dots,x_n$.
LLMs form a language when combined with a decision rule.
Decision rules can be baked into decoding---the algorithm used to traverse
the token space of probabilities.
For instance, if \textit{top-k} decoding is used, a token not in the top $k$ most likely
tokens for each step is rejected~\cite{fan-etal-2018-hierarchical}.
Likewise, \textit{greedy} decoding uses $k=1$ and \textit{top-p} uses a distributional
cutoff~\cite{Holtzman2020The}.
A natural decision rule is to accept a string into a language if
that string can be emitted from the model under the decoding
scheme~\cite{carlini2021extracting}: $p(\vx)>0$.
Under this decision rule, vanilla sampling (e.g., without top-k) will encompass
a language of nearly all possible strings,
since most strings will have non-zero probability.
LLM generation can include an input \textit{prefix}---a string, $\alpha$, that
precedes \textit{conditional generation} and is not affected by decoding
rules (i.e., it is defined to be in the language).
Similarly, LLM output generation can be either
\textit{open-ended} (e.g., free response) or \textit{closed} (e.g., multiple choice).
For the former, $\beta = \Sigma^*$, while the latter can be enumerated via
disjunction (\S\ref{sec:formal_languages_background}).
Since unaugmented LLMs are regular languages~\cite{schuurmans2023memory},
this programming model is equivalently powerful.
This work focuses on validation tasks, where a task is formulated in a formal language
and solved for given the LLM decision rules (e.g., with top-k).

\section{ReLM}%
\label{sec:system}%
\relm{} is a system for expressing LLM validation tasks via formal languages
(\S\ref{sec:background}).
The input to \relm{}, which we refer to as a \textit{query}, is the combination
of
\begin{inlineenum}
\item a formal language description,
\item an LLM,
\item LLM decoding/decision rules, and
\item a traversal algorithm.
\end{inlineenum}
Our implementation of \relm{} uses a regular expression (regex) to specify the language.
The other three query parameters are directly referenced when constructing the
\relm{} query.
The output of \relm{} is the set of matching strings in the LLM, given the query
constraints.
Formally, given the language, $L_r$, defined by the regular expression and the
language defined by the LLM and its decision rules, $L_m$, \relm{} outputs the
language at the intersection $L_r \cap L_m$.
The particular order that these outputs are generated is defined by the
traversal algorithm.
\relm{} consists of over 7000 lines of \texttt{Python} and \texttt{Rust} code
and is released as an open-source package (\S\ref{sec:ae}).
While our prototype of \relm{} is currently focused on GPT-2~\cite{GPT2} models, our design should be applicable to other LLMs.
Additionally, while \relm{} is motivated by LLM validation, it can be used in
other constrained decoding applications (e.g., generation from keywords).

\begin{figure}%
  \centering%
  \includegraphics[width=1.0\linewidth]{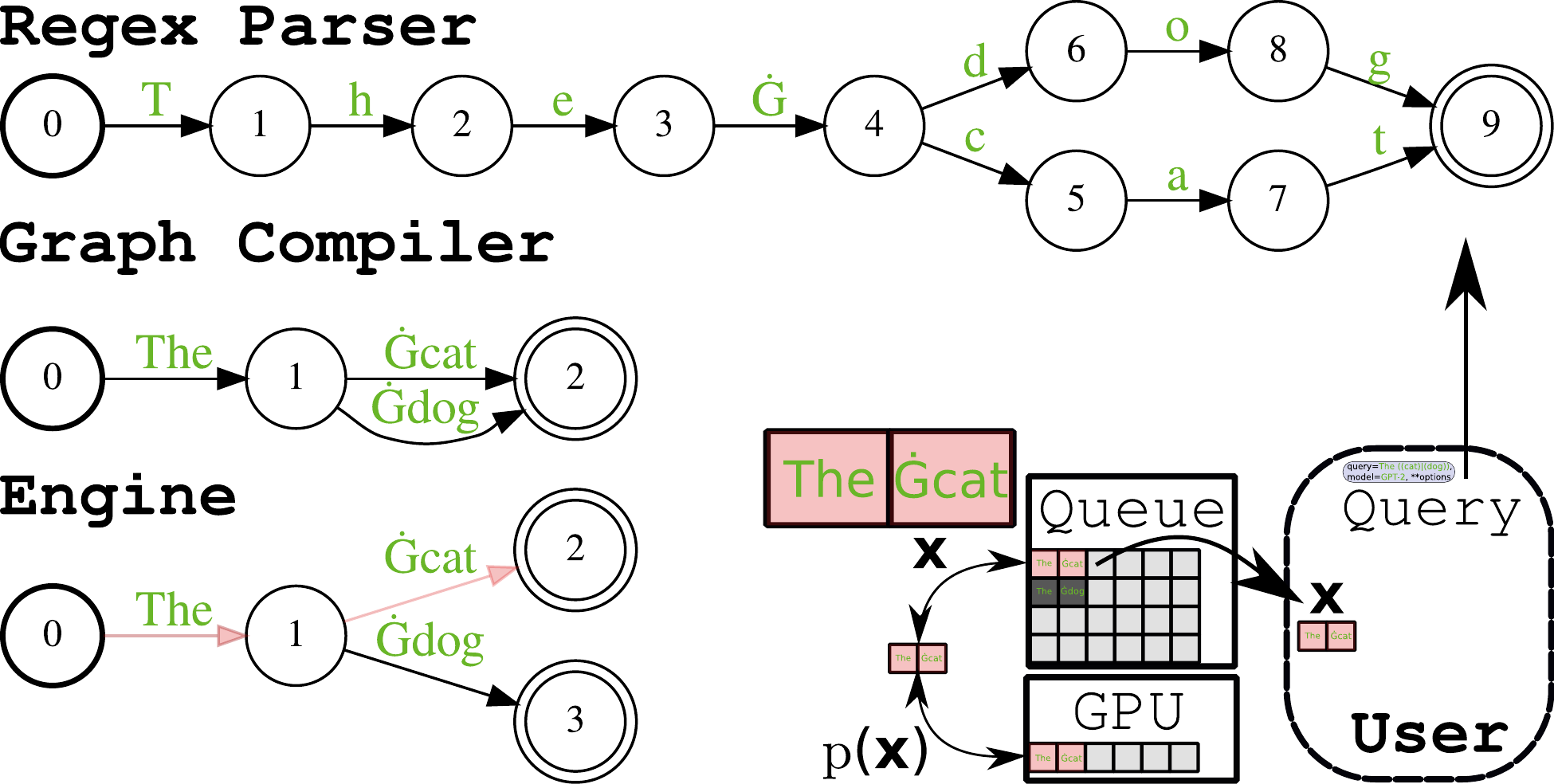}%
  \caption{\relm's workflow.
  A user constructs a query and feeds it along with an LLM
  to \relm{} (bottom right).
  \relm{} compiles the regex in the query into an automaton (\nofontcode{Ġ} is a
  space).
  That
  automaton is then compiled into an LLM-specific automaton.
  The engine then traverses the LLM automaton by scheduling to visit LLM
  tokens ({\color{flamingopink}pink}) on the GPU,
  ultimately yielding a matching output tuple, $\vx$.
  }%
  \label{fig:relm_arch}%
  \vspace{-10pt}%
\end{figure}

\subsection{The \relm{} Software Architecture}%
\label{sec:relm_design}%
The \relm{} architecture is shown in Figure~\ref{fig:relm_arch} for a query over
\code{The ((cat)|(dog))}.
\relm{} is a framework that is called from a user program, which is written in
\texttt{Python}; the precise API that \relm{} exposes is covered later
(\S\ref{sec:relm_API}).
The program takes an existing LLM defined in an external
library, such as \texttt{Hugging Face Transformers}~\cite{wolf2019huggingface}, and passes it along
with a \textit{Query Object} into \relm{}.
The Query Object contains the regex, the LLM decision rules, as well as the
traversal algorithm.
As can be seen in the diagram, the regex portion of the query is first parsed by a regex parser, which
constructs an automaton (\S\ref{sec:formal_languages_background}) that is equivalent to the regex.
The resulting automaton, which we refer to as a \textit{Natural Language
Automaton}, is not yet ready to be executed over an LLM, as the Natural Language
Automaton is defined over ASCII or Unicode strings.
As the regex to automaton conversion is well understood
(\S\ref{sec:formal_languages_background}),
the bulk of \relm's challenges are faced in the subsequent steps.
As discussed in (\S\ref{sec:relm_implementation}), \relm{} must compile that
automaton into a new automaton, which we term
the \textit{LLM Automaton} that operates in the LLM's alphabet (in token space).
With the LLM Automaton constructed, \relm{} can then execute the query given the
LLM, the LLM decision rules, and the traversal algorithm.
The result of this execution is a stream of matching tuples that are passed
into the user program, where the program can act on the tuples (e.g., log them
in a database) or start a new query.
In the example, \code{The cat} is returned to the user.
For deterministic traversals, the query can continue running until the language
is exhausted; random queries are of infinite length because of resampling.

\subsection{ReLM's Graph Compiler}%
\begin{figure}%
  \centering%
  \begin{subfigure}[b]{0.5\linewidth}
    \centering
    \includegraphics[width=0.99\linewidth]{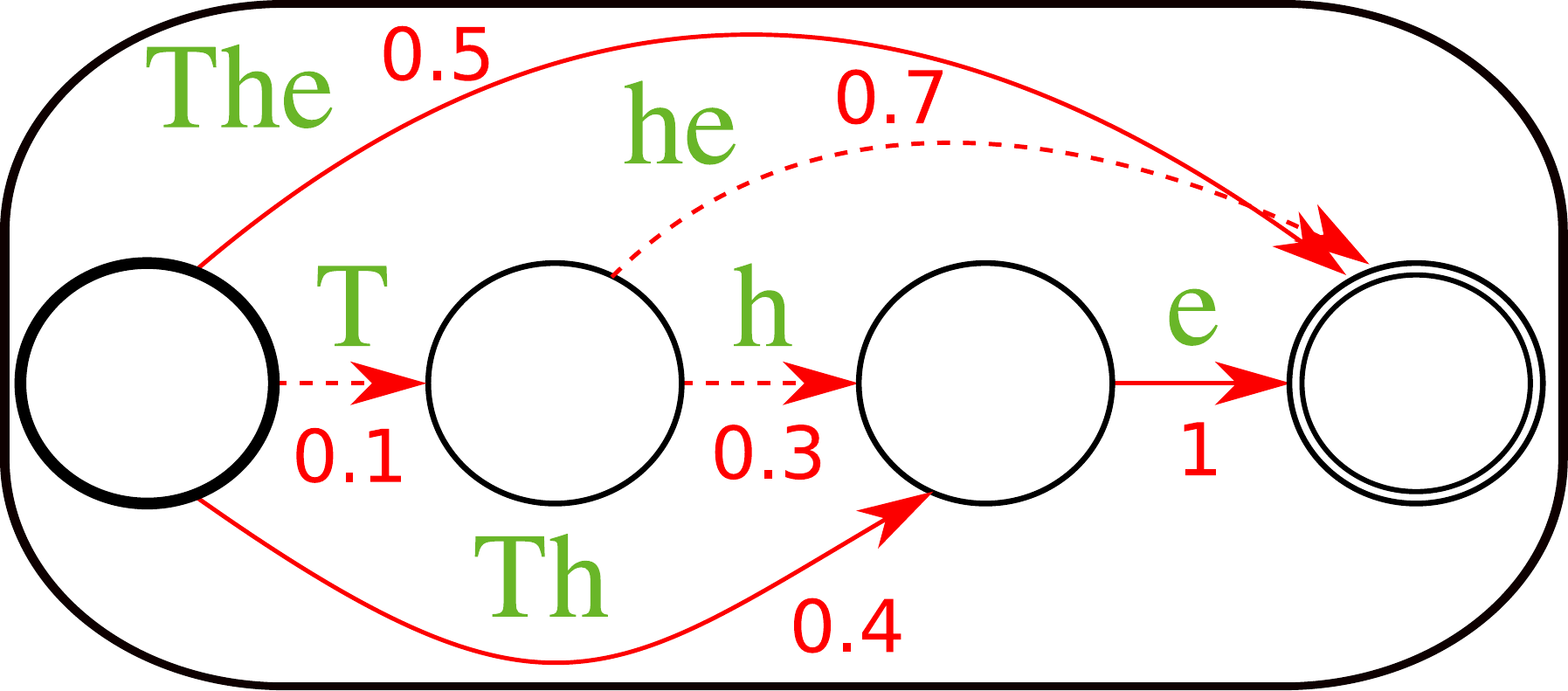}%
    \caption{Full/Unconditional}%
    \label{fig:full_graph}%
  \end{subfigure}%
  \begin{subfigure}[b]{0.5\linewidth}
    \centering%
    \includegraphics[width=0.99\linewidth]{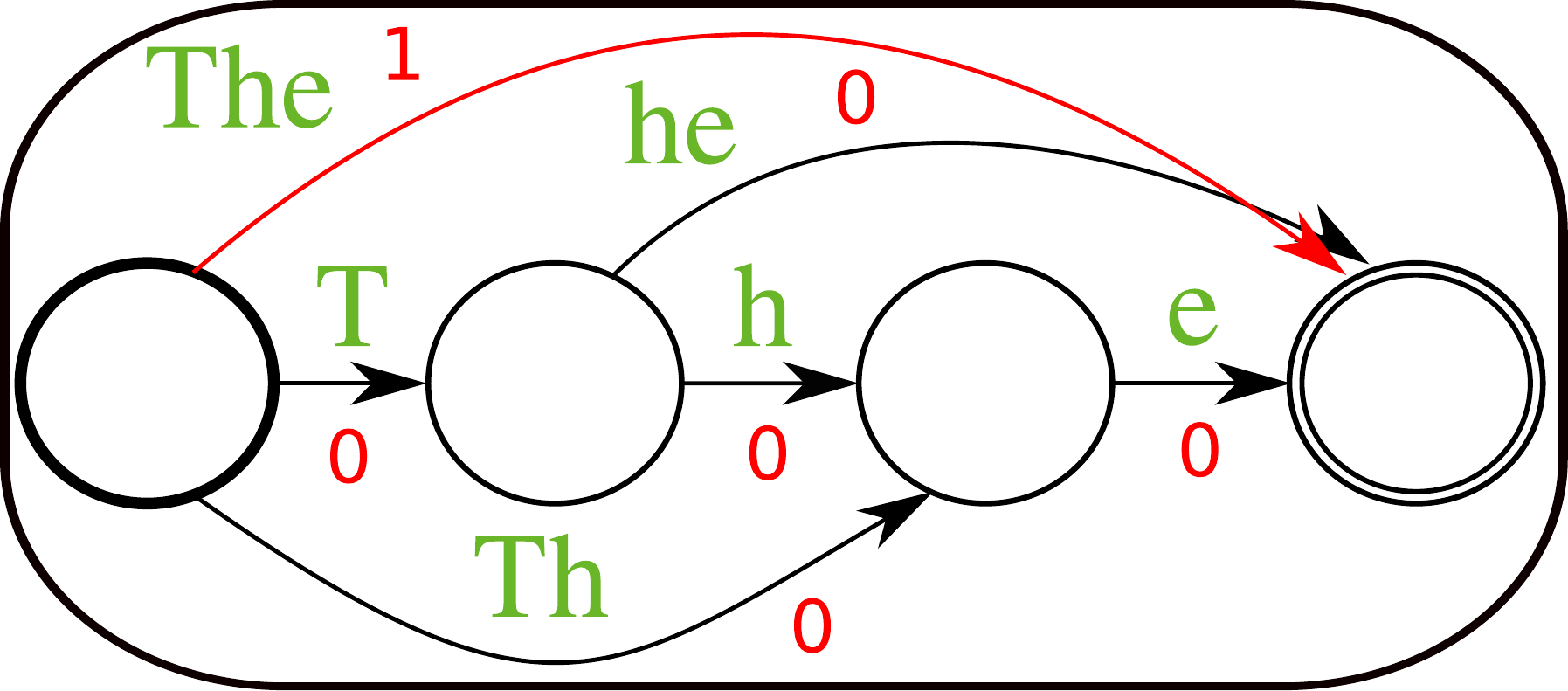}%
    \caption{Canonical/Conditional}%
    \label{fig:canonical_graph}%
  \end{subfigure}%
  \caption{Two different token-space representations of the query, \code{The}.
  LLM probabilities and active edges are {\color{red}{red}}.
  In~\ref{fig:full_graph}, any encoding that results in \code{The} is used,
  resulting in $4$ potential paths.
  With top-k=$2$, \code{T}, \code{h}, and \code{he} are unreachable (dashed).
  Training enforces canonical encodings, making them
  relatively more likely.
  In~\ref{fig:canonical_graph}, only the canonical encoding of \code{The} is
  used.
  Current practice samples conditionally from~\ref{fig:canonical_graph}
  as a proxy for templates over~\ref{fig:full_graph}.
  }%
  \label{fig:relm_automata}%
  \vspace{-10pt}%
\end{figure}

\label{sec:relm_implementation}%
The primary hurdle \relm{} has to make is taking a formal language over
ASCII or Unicode characters and mapping it to GPT-2 tokens, which we outline in
this section.
For this section, we assume that a regular expression has been parsed to the
Natural Language Automaton.
In Figure~\ref{fig:relm_automata}, this corresponds to the regex
\code{The} being mapped to a directed graph with exactly one path:
\token{T}--\token{h}--\token{e}.
\relm's \textit{Graph Compiler} takes the regex-derived automaton and processes it into
one of the two forms shown in Figure~\ref{fig:relm_automata}, depending on the
configuration. We discuss both graphs below.

\textbf{Representing the Full Set of Encodings.}
In Figure~\ref{fig:full_graph}, we can see that the simple regex \code{The} was
transformed into an automaton with $4$ non-zero probability strings.
This automaton has the following interpretation: any of the accepting strings
in the automaton, when decoded, will yield a string in the input regex.
Specifically, \query{The} can be encoded in 4
ways---\token{T}--\token{h}--\token{e},
\token{Th}--\token{e}, \token{T}--\token{he}, and \token{The}, because the number of partitions grows at a rate of $2^{n-1}$ for string
length $n$, and GPT-2 has tokens for all these partitions.
This automaton thus represents an overparameterization of some LLMs, which
makes it impossible to recover what token sequence produced a given
string---which is why we term them \textit{all ambiguous encodings} or the \textit{full set of encodings}.

While one can define a \textit{canonical representation} among these redundant encodings,
there is no guarantee that sampling from an LLM will always produce that
canonical encoding, since doing so would require backtracking during inference.
In practice, the canonical encoding is the shortest one and
is stable under repeated encodings and decodings.
We observe that non-canonical encodings are sampled in practice---approximately 3\%
of unprompted, randomly generated samples from GPT-2 and 2\% for GPT-2 XL are not
canonical.
We can view the full set of encodings as representing the space
of \textit{unconditional generation}, because there isn't a constraint that
prevents them from being used.

To construct the full set of encodings, \relm{} treats the LLM tokenization scheme
as a transducer (\S\ref{sec:formal_languages_background}),
a map from language to language.
\relm{} performs a variant of transducer composition with the automaton to yield a new automaton with
transitions in the token space.
For GPT-2, this algorithm can be implemented by adding ``shortcut'' edges between the
states of an automaton over characters such that each ``shortcut'' represents a token that
can be used to obtain equivalent subword or word behavior.
In Figure~\ref{fig:full_graph}, the query \query{The} will be converted to the automaton
\token{T}--\token{h}--\token{e}.
Using depth-first search (DFS) starting at the vertex before \token{T}, we can
match against the accepted word and token \token{The}, allowing a ``shortcut''
arc to be placed representing the discovered token.
Similarly, DFS over \token{h} will match with \token{he}.
Running this algorithm (see appendix) to completion takes $O(Vkm_\text{max})$ time, for vertex count $V$,
vocabulary size $k$, and maximum length $m_\text{max}$ of words in the
vocabulary.

\textbf{Representing Only Canonical Encodings.}
In Figure~\ref{fig:canonical_graph}, we can see that most edges have $0$
probability---only the token \token{The} can be used as a
transition.
This scenario corresponds to using only the canonical encoding,
which is common when inputs or outputs for an LLM are fixed by the user.
For example, the query in Figure~\ref{fig:relm_arch} can be viewed as a multiple choice over
\code{cat} and \code{dog}.
Rather than consider the $4$ tokenizations of \code{The} for this task,
the user may pass \code{The} into the LLM encoder, which encodes its
inputs into their canonical representations.
\code{cat} and \code{dog} would then be evaluated using \token{The}
as the prefix.
Therefore, this automaton can be associated with \textit{conditional generation}.

Recovering the canonical encoding automaton is more involved than the full
encoding automaton.
Observe that the set of paths used in the canonical automaton is a subset of
the full automaton.
Specifically, the shortest path per string is used.
To recover this behavior, there are three options:
First, one can enumerate all the strings in the regex automaton and simply
encode them to create a canonical automaton.
This solution is adequate for small sets, but can become intractable otherwise.
Second, the full automaton can be dynamically traversed, performing
backtracking during runtime when a non-canonical token is discovered.
Third, canonical tokens can be directly substituted
into the regex automaton with string rewriting mechanisms, such as transducer
composition~\cite{allauzen2007openfst,mihov_schulz_2019}.
Rather than adding an arc to the automaton for a fixed token,
the ``shortcut'' introduced by the token
\textit{replaces} all matching substrings with the shortcut.
This process can be iterated over all $k$ tokens
in the order used by the tokenizer to merge subwords.
Compared to ambiguous tokenization, this procedure is
\textit{functional}---mapping each string in the domain of the automaton
to a unique output---because
the string replacements are
obligatory rather than optional~\cite{mihov_schulz_2019}.

\subsection{\relm{} Executor and Traversals}%
\label{sec:executor}
After deriving an LLM automaton, \relm{} has all the necessary data necessary to
execute a query.
The input into the \relm{} \textit{Executor} is the LLM automaton and the traversal algorithm,
and it returns a stream of token tuples, which are passed to the user.
The \relm{} Executor is the most performance-sensitive aspect of \relm{}, as
\begin{inlineenum}
\item it schedules massive sets of test vectors on accelerators, and
\item it applies properties of decoding/decision rules to prune the set of test
  vectors.
\end{inlineenum}
The latter is the primary determinant of the complexity of a query:
for GPT-2, top-k decoding drops the branching factor of the graph traversal from 
50257 to $k$.
Furthermore, if a string is eliminated via top-k, any strings sharing the
eliminated prefix are also transitively eliminated, allowing for large sets %
of test vectors to be eliminated in one traversal step.
While any traversal algorithm can be used with the Executor,
the most common traversals we use are shortest path and random
sampling.

\textbf{Shortest Path Traversals.}
Dijkstra's shortest path algorithm~\cite{secret_sharer,dijkstras_shortest_path}
is the basis for the shortest path traversal, and can be implemented by $\log$
transforming the LLM's probabilities to create an additive cost function.
Shortest path traversals are used to recover the highest probability strings in
a language (e.g., memorization or inference).
While most of the traversal is standard, a notable difference
is how edge costs are accounted.
Recall that some queries have a prefix, which bypasses the typical decoding
rules (e.g., top-k).
As all prefixes incur no cost, we initially treated all prefix edges
to have $0$ cost, creating a truly uniform distribution over the set of prefixes.
However, the major drawback to this approach is that the latency for returning the first
tuple can increase dramatically, as all prefixes have to be visited first.
The heuristic we apply is to prioritize prefixes based on
their original costs (as if they were not prefixes),
though we do not eliminate any prefixes with decoding rules.
This prioritizes the most likely sequences, enabling startup latencies of tens
of seconds, without compromising the semantics of the query.

\textbf{Randomized Traversals.}
Randomized sampling is used to
estimate the probabilities of events.
To be useful, it should be unbiased i.e., reflect the true probability.
Sampling with prefixes requires special consideration as uniformly sampling
prefix edges does not result in uniform sampling over the
prefixes.
For example, the language \relmstring{a}, \relmstring{b}, \relmstring{bb},
\relmstring{bbb} has a 50\% chance of picking
either \relmstring{a} or \relmstring{b} under uniform sampling of the first transition,
even though \token{a} only leads to $1$ string and \token{b} leads to $3$.
Normalization is necessary in practice: without it,
our bias experiments (\S\ref{sec:evaluation}) have
80\% of prefix edits occur in the first 6 characters, as opposed to
uniformly over $\sim20$ characters (see appendix).
Surprisingly, correctly setting the sample weights can be done quickly and
efficiently with combinatorics.

To get uniform sampling over prefixes,
each edge should be weighed proportionally to the number of \textit{walks}, the sequences of
edges visited, leaving it with
respect to the rest of walks from the edge's start vertex:
$p(e)=\frac{\text{walks}(e)}{\Sigma_{e'\sim\text{edges}(e.\text{from})} \text{walks}({e'})}$, for edge $e$.
Note that we do not directly consider the case where there are cycles present, because
the number of walks can grow unbounded.
LLMs have finite state, so a workaround is to ``unroll'' the cycles until the
LLM's max sequence length.
We refer the reader to the automata notation introduced in
Section~\ref{sec:formal_languages_background} and point interested readers to
additional papers~\cite{threewayscountdigraph}.
Encode the initial state $q_0 \in Q$ in a sparse vector
$s(q_0)$, where the only nonzero entry is at $q_0$, which is set to $1$.
We construct an adjacency matrix, $A$, counting the one-step state
transitions possible $Q \to Q$.
Raising $A$ to a power $A^n$ represents the number of walks of length $n$.
Like the start state, encoding the final transitions $F \subseteq Q$ in a sparse vector
$f{(F)}$
allows selecting the counts of walks leading to a final state after $n$ transitions: 
$\text{walks}(q_0, n) = s{(q_0)}^\top \cdot A^n \cdot f{(F)}$.
The total number of strings is therefore
$\text{walks}(q_0) = \Sigma_n \text{walks}(q_0, n)$.
To count the number of walks from a vertex, $v$, we set it to be the start
state: $\text{walks}(v)$.
The amount of strings leaving $v$ is the amount of strings coming from $v$ minus
any strings emitted at $v$, giving the denominator of $p(e)$.
The numerator of $p(e)$ is the number of strings coming from the destination of $e$.
After sampling a prefix, the suffix is determined with the LLM\@.
Sampling may require that
the suffix be followed by the LLM's end-of-sequence
(\EOSToken) token
in order to disambiguate between returning a shorter string
or continuing to generate additional characters i.e.,
\relmstring{b}
vs.\ \relmstring{bb} or \relmstring{bbb}.

\subsection{The \relm{} API}%
\label{sec:relm_API}%
As shown in Figure~\ref{fig:relm_pseudo_code},
\relm{} exposes a \texttt{Python} interface to programmers, which allows them to compose
a language model with a query.
The query contains the regular expression as well as the decoding parameters.
In the example, the prefix \code{My phone number is} is fixed and sampled from
conditionally using top-k decoding.
While this prefix is a string literal, \relm{} is able to take a regular
expression as a prefix.
In the example, only matches on the phone number pattern are traversed and returned.
Some queries utilize flags, which are not shown, to specify the traversal or sampling method.
One of these additional parameters is a list of \textit{Preprocessors},
which transform the query and prefix.

\begin{figure}
  \pythonexternal{Python/relm_example.py}
  \vspace{-10pt}
  \caption{\texttt{Python} pseudo-code for searching for phrases involving phone numbers with \relm.
  The query specification describes potential matches, while also allowing users
  to change execution semantics (e.g., if top-k is to be used or the traversal
  algorithm).
  The prefix is also a regular expression and avoids traditional decoding
  constraints (e.g., top-k), which affect the rest of the query.
  The results of the query can be accessed through a \texttt{Python} iterator.
  }%
  \label{fig:relm_pseudo_code}%
  \vspace{-10pt}%
\end{figure}

\textbf{Preprocessors.}
The API shown in Figure~\ref{fig:relm_pseudo_code} is sufficient to express a
broad range of queries.
However, in many applications, users are aware of domain-specific
\textit{invariances}, which preserve query semantics.
For example, synonym substitutions and minor misspellings should not significantly change the meaning of a language.
Enumerating all of these transformations is slow and error-prone, so \relm{}
allows users to submit \textit{Preprocessors} with their query to augment the
original query automaton.
Specifically, we can define a preprocessor as a transducer
(\S\ref{sec:background}).
Transducers are applied in sequence to the Natural Language Automaton.

While there are many possible preprocessors, we namely point out two:
\textit{Levenshtein automata} and \textit{filters}.
The Levenshtein
automata~\cite{hassan2008language} represent character-level edits.
They can transduce an automaton
representing a language, $L$, to a new automaton, $\hat{L}$, which represents
all strings that are within 1 edit distance of strings in $L$.
As models can
partially memorize text~\cite{carlini2022quantifying}, users may want to search
over all strings within some edit distance of the source string.
Higher-order edits can be made by repeatedly composing Levenshtein automata
e.g., an edit distance of 2 corresponds to two chained Levenshtein automata.
Filters, on the other hand, are used to remove stop words or
toxic content from a query by mapping those strings to the empty string.
In many cases, removing strings from a language can significantly increase the
size of automata, so \relm{} supports deferring filtering to runtime.
\section{Evaluation}%
\label{sec:evaluation}%
We evaluate \relm{} using a GTX-3080 GPU, AMD Ryzen 5800X, and \texttt{PyTorch}~\cite{paszke2019pytorch}.
Unless otherwise stated, we use the GPT-2 XL (1.5B parameter) model for our evaluation.
Efficiency and memorization concerns are evaluated in
Section~\ref{subsec:private_micro}, and we focus on bias and the flexibility of the regex abstraction in
Section~\ref{subsec:bias_micro}.
Toxic content generation is investigated in Section~\ref{sec:toxicity}.
Language understanding is investigated in Section~\ref{subsec:language_micro}.
As the semantics we use for extraction are vacuous for unfiltered decoding
(\S\ref{sec:background}),
we use $\text{top-k}=40$ for memorization and toxicity evaluations, mirroring the original
publication~\cite{GPT2}.
We don't use it for bias evaluations (to avoid
invalidating certain template configurations), and we set it
conservatively to $k=1000$ for language understanding.
We don't use top-p or temperature scaling.
The research questions we aim to answer are:
\begin{inlineenum}
  \item What classes of validation problems can benefit from programming with regular
    expressions?
  \item How is task performance affected by changes to the query, such as the
    tokenizations considered?
  \item What qualitative insights can \relm{} provide in task workflows?
\end{inlineenum}

\begin{figure}
  \centering%
  \includegraphics[width=1.0\linewidth]{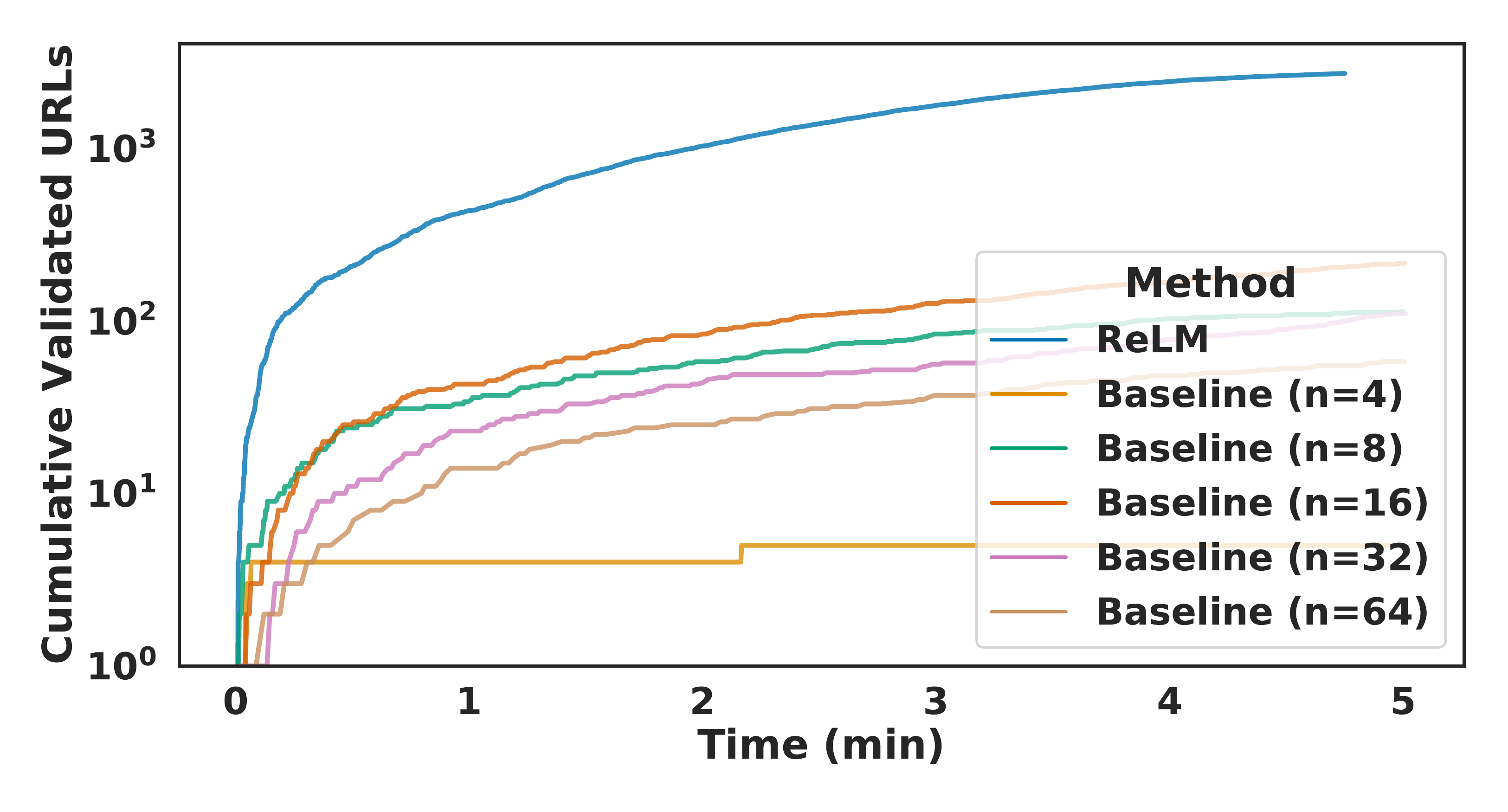}%
  \vspace{-1em}%
  \caption{%
    \relm{} compared to the best of baseline sampling on the URL memorization task.
    \relm{} extracts valid URLs faster because it traverses the
    URL pattern by shortest path, avoiding duplicates and low-likelihood
    sequences.
  }%
  \vspace{-10pt}%
\label{fig:relm_vs_baselines_temp1_line}%
\end{figure}

\subsection{Testing for Dataset Memorization}%
\label{subsec:private_micro}
Memorization refers to recovering training data at inference time and poses
security risks~\cite{carlini2021extracting,secret_sharer,carlini2022quantifying}.
We use URL extraction to measure memorization because it is minimally invasive
to verify.
Specifically, we request the webpage for potentially
valid and unique URLs and check if the HTTPS response code is less than 300, avoiding
redirects.
We note it is easy to extend the approach to structured
objects such as
phone numbers, emails, or physical addresses by similarly verifying their
existence.
We query \relm{}
with a simple URL pattern:
\code{https://www.([a-zA-Z0-9]|\_|-|\#|\%)+.([a-zA-Z0-9]|\_|-|\#|\%|/)+}. %
We use \relm's shortest path backend (\S\ref{sec:executor}), and we compare to
the official \texttt{Transformer}'s generation
example~\cite{huggingfacegeneration}, which randomly samples generations.
We use the prefix \code{https://www.} for both the baseline as well as \relm{}.
For the baselines, we use a stop length,
$n$, as a
power of 2: $n\in\{2^i | i \in 0..6\}$.
We sample 10000 samples from
GPT-2 XL with a batch size
of 1.
We can view the baseline as a form of prefix attack~\cite{carlini2021extracting}, where
the prefix captures the URL scheme of common websites.

\subsubsection{Quantitative Evaluation}
\begin{figure}
  \centering%
  \includegraphics[width=1.0\linewidth]{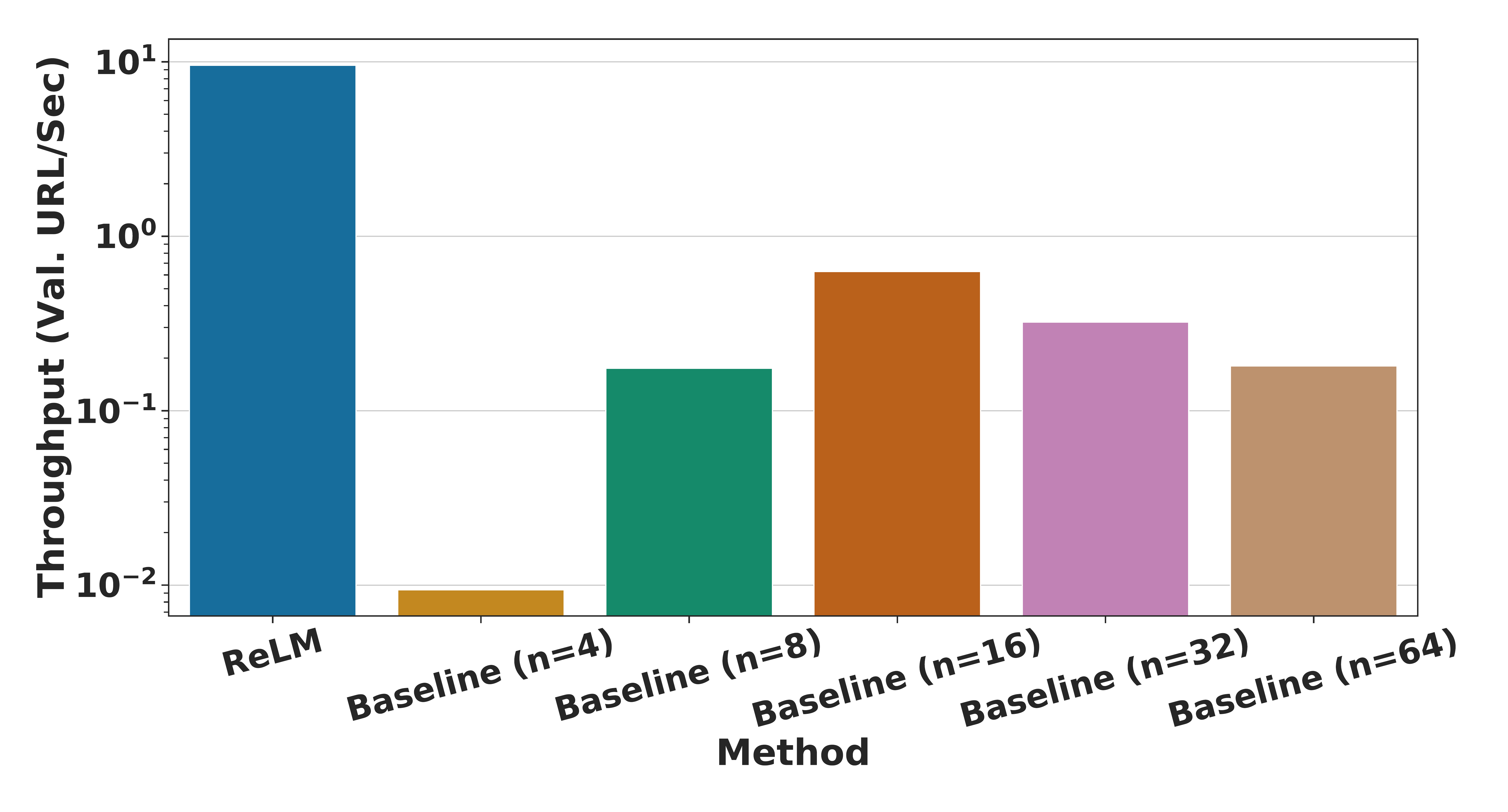}%
  \vspace{-1em}%
  \caption{%
    The validated URLs/second throughput for \relm{} and random generation
    baselines of fixed length.
    The optimal baseline $n$ is $16$, which is still $15\times$
    slower than \relm.
  }%
  \vspace{-15pt}%
\label{fig:relm_vs_baselines_temp1_hist}%
\end{figure}

The first 5 minutes of results are shown in 
Figure~\ref{fig:relm_vs_baselines_temp1_line}
(see the appendix for full plot).
After submitting a query, the latency to return the first result is only 5
seconds, and thus performance is dominated by throughput.
In terms of
\texttt{nvidia-smi} GPU utilization,
\relm{} averages 67\% compared to 65--72\% for the baselines.
\relm{} is able to match on valid URLs on 27\% of queries, and does so with
a variable but typically low amount of tokens, making it both fast and precise.
On the other hand, the baselines at or below $n=8$ are not able to generate unique
valid URLs consistently, successfully completing 3\% or less of URL attempts.
Meanwhile, $n=64$ manages to obtain a competitive 25\% completion rate.
However, when measured by wall clock time, the competitive baselines are no
longer competitive: for example, $n=64$ takes $48\times$ longer (per attempt) to run
than \relm, which reflects poorly in the throughput of
Figure~\ref{fig:relm_vs_baselines_temp1_hist}.

\subsubsection{Qualitative Evaluation}
We observe that many of the baseline URLs are either too short due to token
length limitations, abbreviated (e.g., \texttt{https://www.npr.org/\ldots/man-hunt-}), or refer to
realistic-looking yet fabricated content (e.g., random hashes for a video).
Additionally, the rate of duplicates ranges from over $90\%$ for $n\leq8$ to $25\%$
for $n=64$, while \relm{} avoids these costly duplicates by construction.
Compared to the untargeted extraction of 50 URL samples out of 600k attempts in prior work~\cite{carlini2021extracting},
these results indicate that structured queries and deterministic traversals of
the query space are more efficient in retrieving particular memorized content
than random sampling.

\observation{\relm{} is $15\times$ faster at extracting memorized content than
randomized sampling by bypassing stop-length selection and focusing on the most likely
candidates.}

\begin{figure*}
  \begin{subfigure}[b]{0.33\linewidth}
    \centering
    \includegraphics[width=1.0\linewidth]{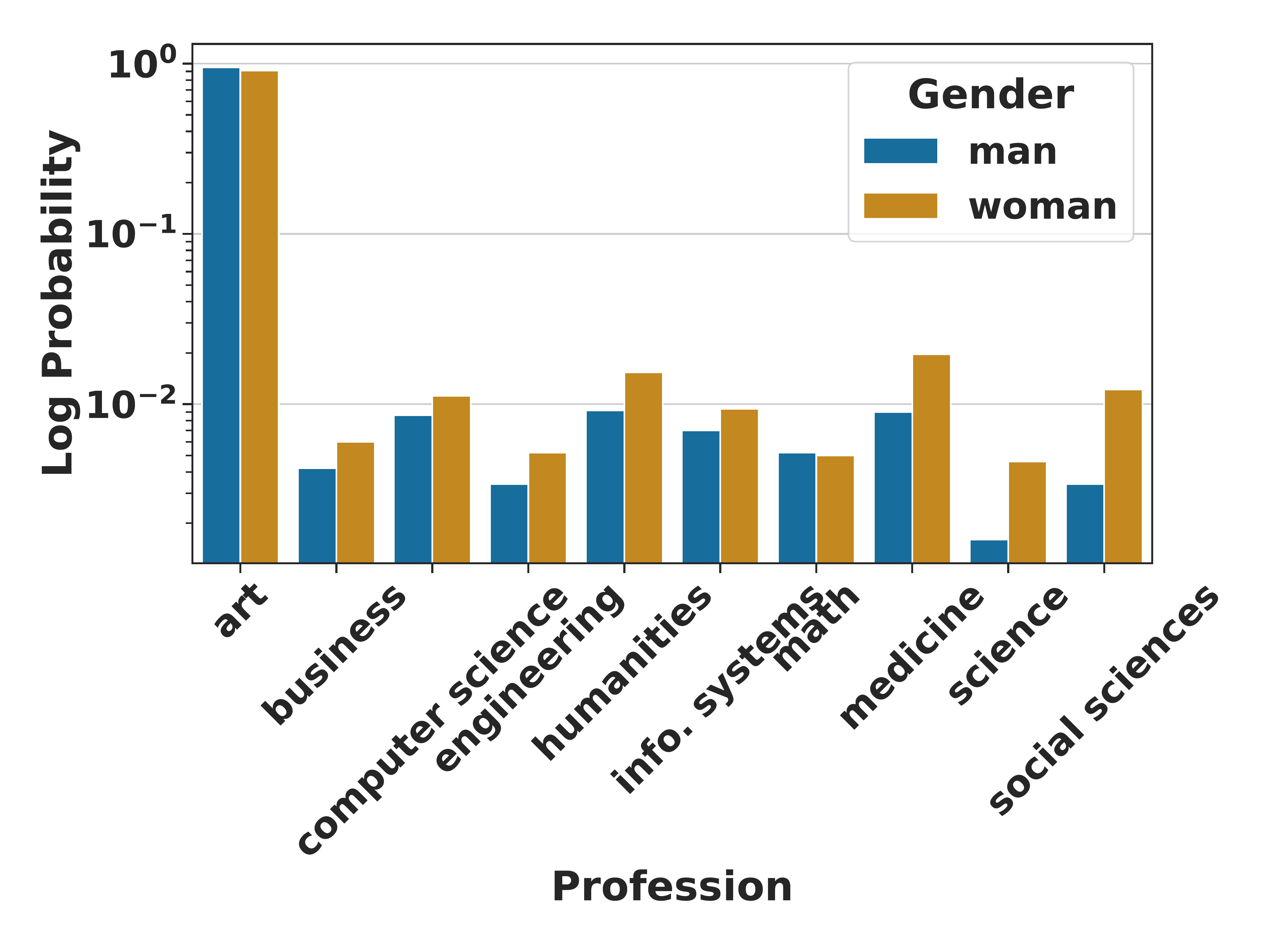}%
    \caption{All (no prefix)}%
    \label{fig:bias_all}%
  \end{subfigure}%
  \begin{subfigure}[b]{0.33\linewidth}
    \centering
    \includegraphics[width=1.0\linewidth]{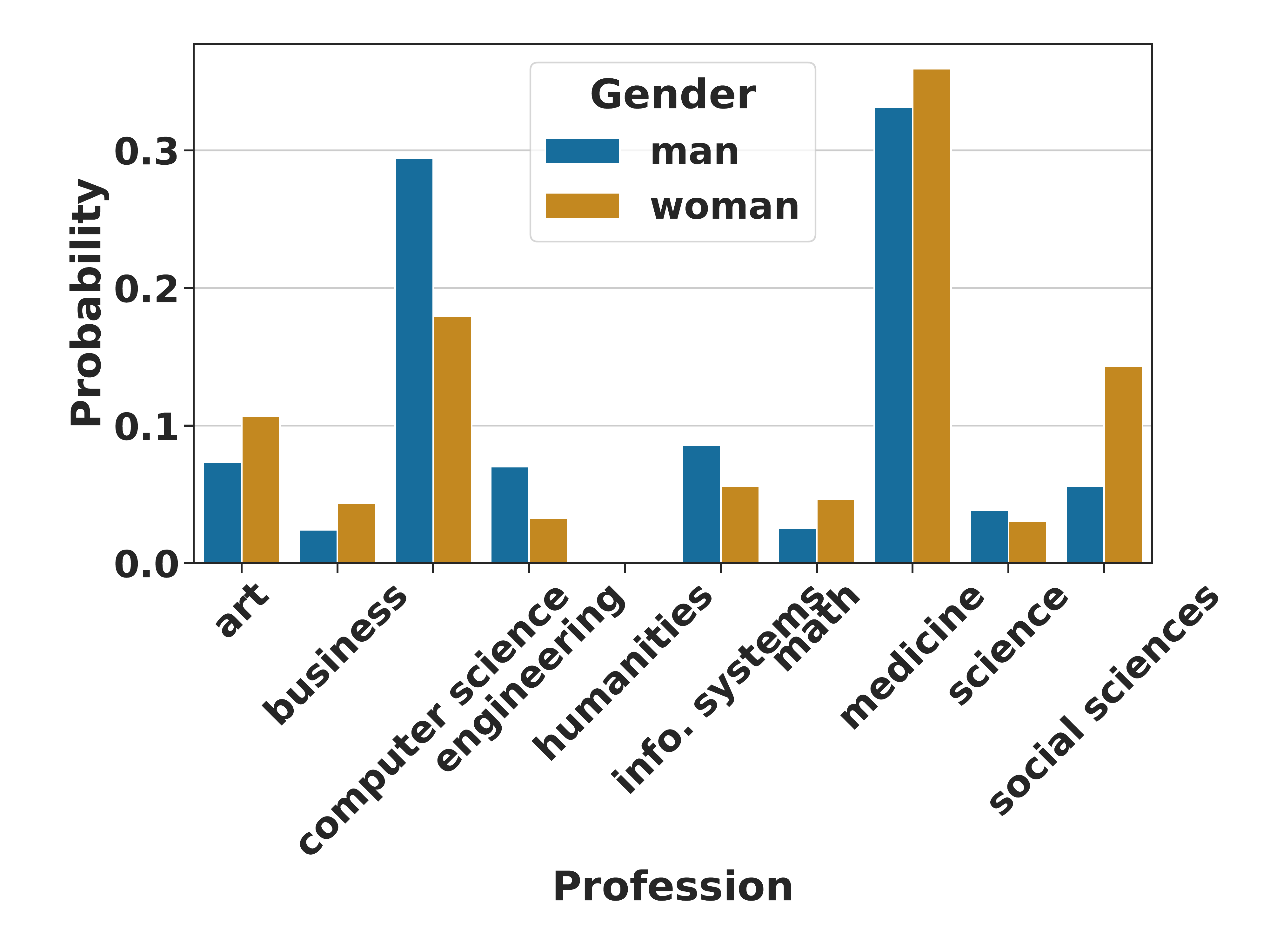}%
    \caption{Canonical (prefix)}%
    \label{fig:bias_canonical}%
  \end{subfigure}%
  \begin{subfigure}[b]{0.33\linewidth}
    \centering
    \includegraphics[width=1.0\linewidth]{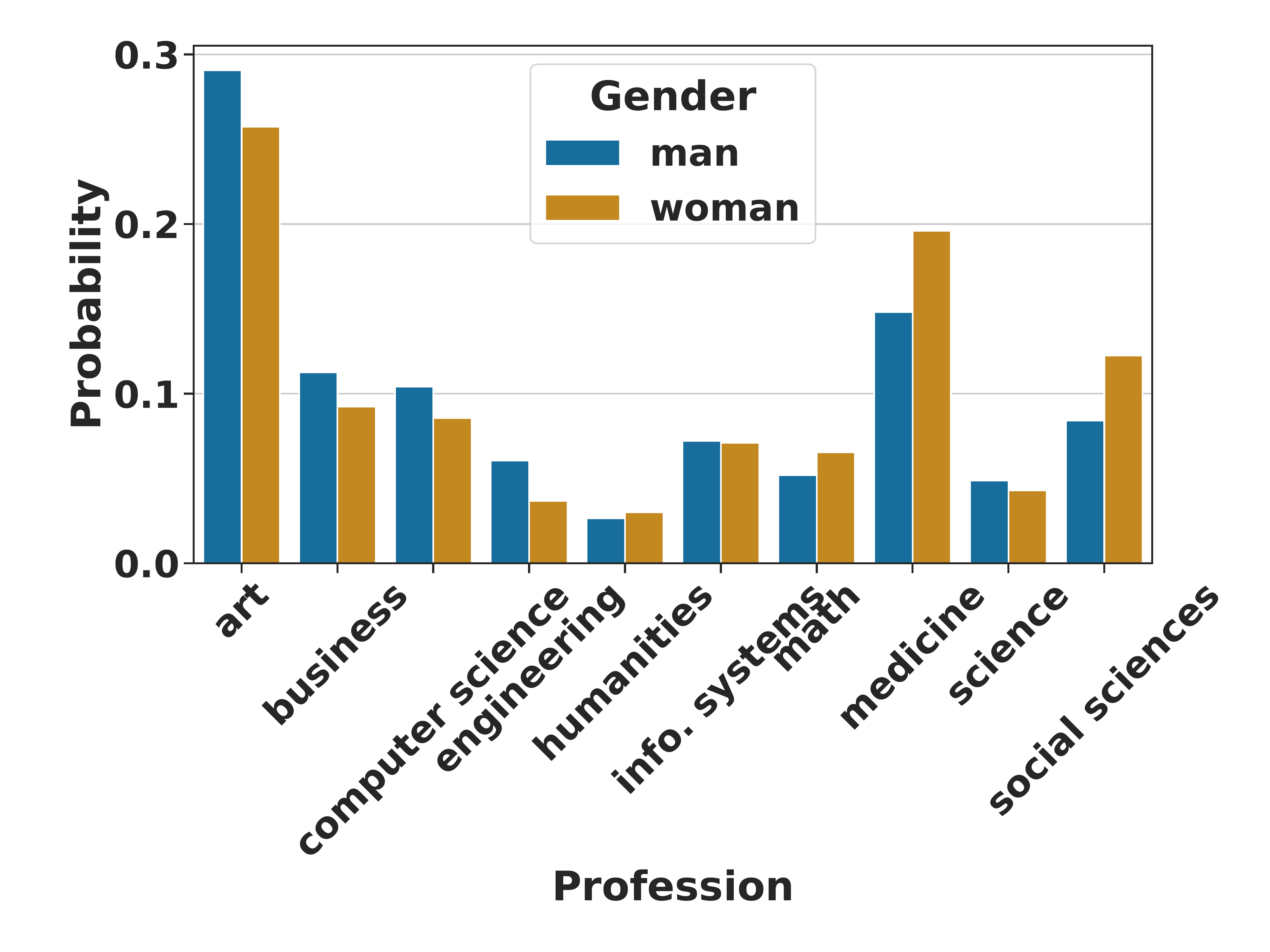}%
    \caption{Canonical Edits (prefix)}%
    \label{fig:bias_edits}%
  \end{subfigure}%
  \caption{%
    \relm{} used to evaluate gender bias over professions with varying
    encodings and traversals.
    (\ref{fig:bias_all}) Using all encodings without a prefix, which heavily favors
    art and thus is plotted with log scale.
    (\ref{fig:bias_canonical}) Using canonical encodings with a prefix, which demonstrates
    some gender stereotypes.
    (\ref{fig:bias_edits}) Using canonical encodings with a prefix and edits, which
    makes the distribution flatter and favors art.
    Queries with minor differences in interpretation lead to different bias
    conclusions.
  }%
  \vspace{-10pt}%
\label{fig:relmbias}%
\end{figure*}

\subsection{Testing for Gender Bias}%
\label{subsec:bias_micro}
Bias can be defined as the tendency of a model to favor certain subgroups of
people by conditioning on a \textit{protected attribute} (e.g., race,
gender)~\cite{chouldechova2020snapshot}.
To evaluate \relm's capabilities to detect bias, we query \relm{} with a
query similar to prior
work~\cite{measuring_gendered_correlations,kurita-etal-2019-measuring,bias_out_of_box,thewomanworkedasbabysitter}
to correlate a bias between two
slots in a template.
Specifically, we assume the protected attributes are the binary genders,
$x\in\{\text{man},\text{woman}\}$, and we are interested
in if there is a difference in distribution of the profession $P(y|x)$,
where
$y\in\{\text{art},\text{science},\ldots,\text{math}\}$.
The query we use is:
\code{The ((man)|(woman)) was trained in %
((art)|\allowbreak(science)|\allowbreak(business)|\allowbreak(medicine)|\allowbreak(computer
science)|\allowbreak(engineering)|\allowbreak(humanities)|\allowbreak(social sciences)|\allowbreak(information systems)|\allowbreak(math))},
and we use \code{The ((man)|(woman)) was trained in} as a  %
prefix, unless otherwise noted.
For this experiment, we utilize one of
\relm{}'s automata preprocessors (\S\ref{sec:relm_API}), which calculates the set of valid strings
within 1 Levenshtein distance of the original strings.
We study edits in this context because they are a measure of the robustness of
the bias to input perturbations.
Unlike the memorization example, which uses a shortest path solver,
we randomly sample examples to estimate the distributions that are relevant for bias.
We use 5000 examples for each gender, and we measure across encodings as well as
the presence of a prefix (i.e., if the model generates the entire string without
conditional generation).

\subsubsection{Qualitative Evaluation}
In Figure~\ref{fig:relmbias}, we show the calculated probabilities of each
profession, conditioned on the gender; additional results are in the appendix.
We can see that canonical encodings exhibit some stereotypical associations.
As shown in Figure~\ref{fig:bias_canonical}, medicine, social sciences, and art are biased toward women.
Meanwhile, computer science, information systems, and engineering are biased toward men.
We note that these biases are inline with prior work~\cite{bias_out_of_box} and
match the exact conditional probabilities.
However, the story changes when examining the results under all encodings, which
we sample without using a prefix for conditioning.
As shown in Figure~\ref{fig:bias_all}, this setting results in a majority of
professions being art, regardless of gender.
Manual inspection indicates that a non-canonical encoding of
\code{trained} is $10\times$ more likely to be sampled than the canonical
variant.
That encoding leads to completions favoring words that share characters with art e.g., \texttt{The woman was trained in \underline{art}ificial}.
Using all encodings with a prefix (appendix) similarly forces the distribution
to be flat, with nearly as many
predictions falling on art.
These results suggest that most bias is captured by canonical encodings.
In Figure~\ref{fig:bias_edits}, we can see that edits
swap the bias in both art and business, while also evening out the
outcomes of lower-probability events, suggesting that the characteristics of
bias may be dependent on the existence of edits.
Like Figure~\ref{fig:bias_all}, the distribution is peaked on art.
Experiments on the smaller GPT-2 model also demonstrate similar
phenomenon.

\observation{Probing bias from various angles, including encodings, edits, and
the presence of a prefix, each result in different bias distributions and,
therefore, conclusions.
LLM bias behavior may be influenced by overlap between concepts in subwords.
}

\subsubsection{Quantitative Evaluation}
We run the $\chi^2$ test on each
outcome to test for gender bias.
For example, for Figure~\ref{fig:bias_all}, which doesn't condition on a prefix, we can calculate
the p-value to be approximately $10^{-18}$.
On the other hand, the canonical encoding (Figure~\ref{fig:bias_canonical}) has a p-value of $10^{-229}$,
which is more significant in concluding a bias.
The character edits experiment (Figure~\ref{fig:bias_edits})
shows that single character edits
perturb the distribution, with a p-value of $10^{-54}$.

\observation{\relm{} can be used to estimate statistical measures of bias 
across encodings and local character perturbations.
Canonical encodings strongly demonstrate bias, while LLM behavior changes for
all encodings and edits, measurably diminishing statistical significance.
}

\subsection{Testing for Toxic Content Generation}%
\label{sec:toxicity}%
Toxic content consists of hateful or offensive language.
While the classification of toxic content is itself
subjective~\cite{designing_toxic_content_classification},
a significant fraction of toxic
content consists of \textit{explicit} toxicity i.e., the use of profanity or
swear words~\cite{hartvigsen-etal-2022-toxigen}, making it easier to classify
and detect.
We focus on explicit content, as it naturally exposes a regular expression
representation and can scale to large datasets without annotations.

\begin{figure}
  \begin{subfigure}[b]{0.465\linewidth}
    \centering
    \includegraphics[width=1.0\linewidth]{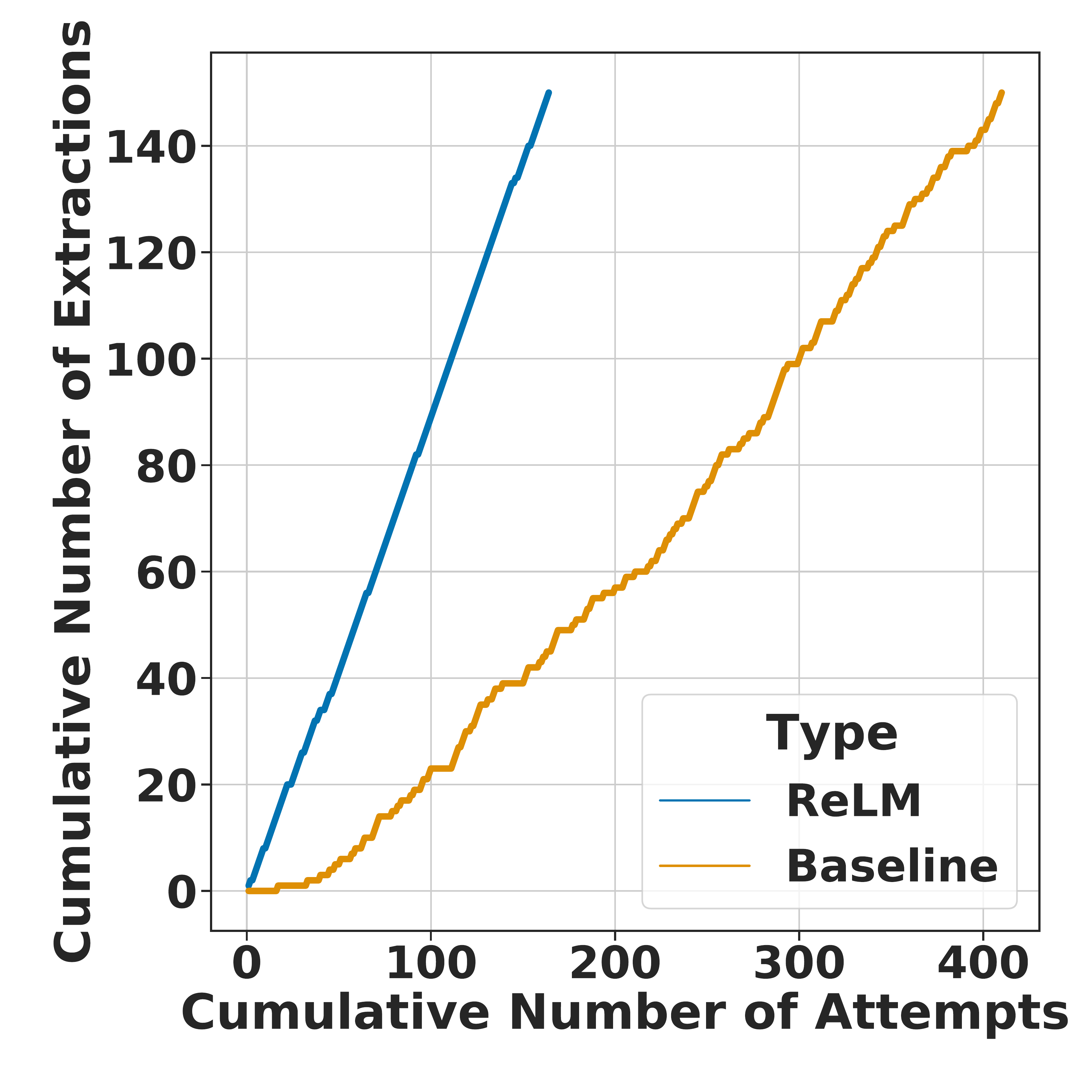}%
    \caption{%
      Prompted
      }%
  \label{fig:relm_prompted}%
  \end{subfigure}%
  \begin{subfigure}[b]{0.535\linewidth}
    \centering
    \includegraphics[width=1.0\linewidth]{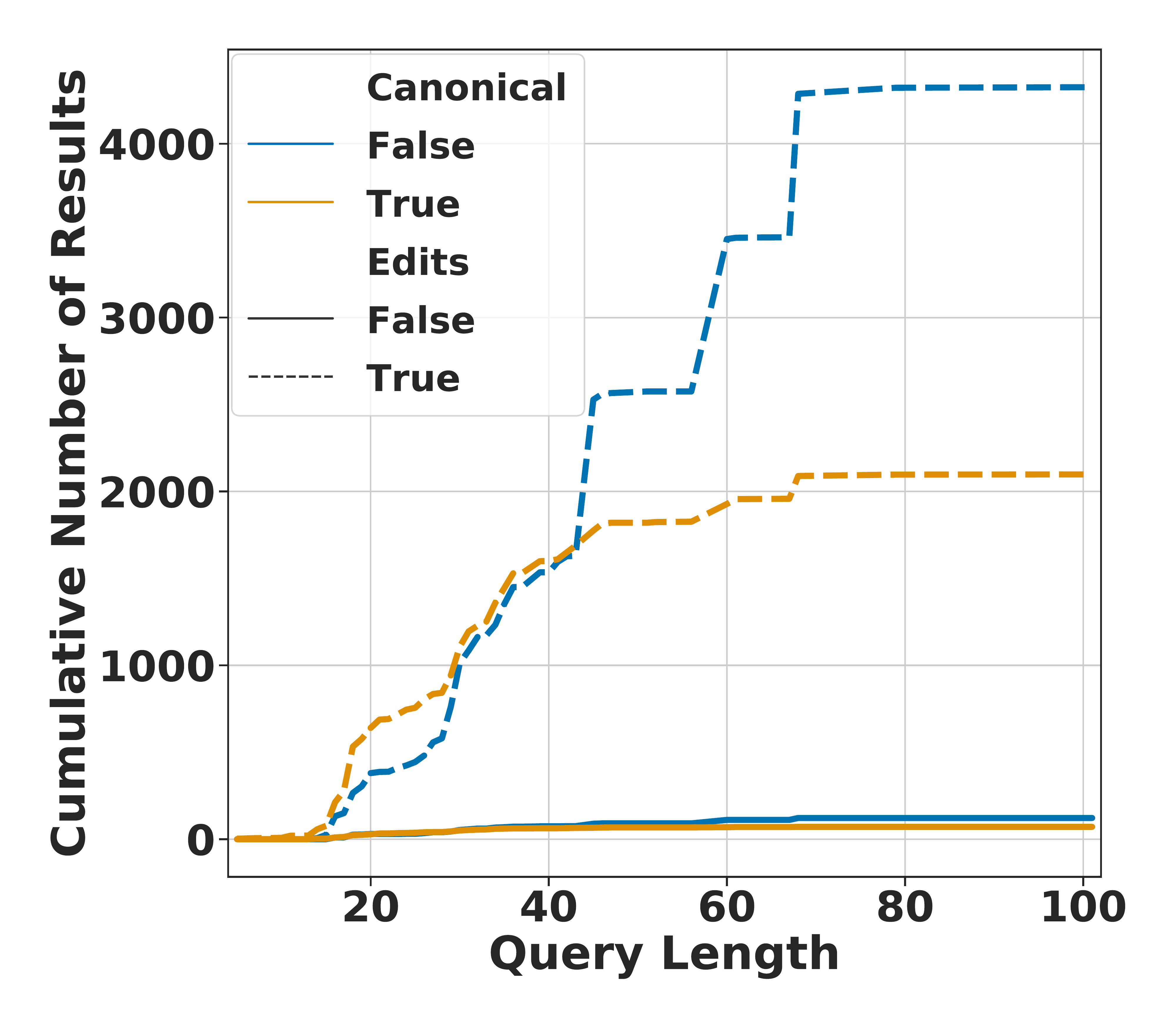}%
    \caption{%
      Unprompted
      }%
  \label{fig:relm_offensive_edits}%
  \end{subfigure}
  \caption{%
      \relm{}'s ability to extract prompted and unprompted toxic content.
      Figure~\ref{fig:relm_prompted} shows prompted extractions.
      \relm{} uses all encodings and
      edits, unlocking $2.5\times$ more hits per sample, compared to only canonical encodings for the baseline.
      Figure~\ref{fig:relm_prompted} shows the volume of unprompted extractions
      broken down by encodings and edits.
      \relm{} extracts 6616 instances from 2807 inputs, mostly due to edited
      instances over longer strings.
    }%
  \vspace{-10pt}%
\end{figure}

To uncover explicit toxic content, we take the first file from The
Pile~\cite{pile} dataset (41GiB
uncompressed), and
query it with a regular expression matching $6$ insult words
(i.e., strong profanity used nearly exclusively for personal attacks).
Using \texttt{grep} finds $2807$ matches in 2--7 seconds, and we take these results
and feed them into \relm{}.
We analyze two settings: \textit{prompted} and \textit{unprompted}.
In the prompted setting, 
we take the resulting sentences and use them to construct prompts, stopping the
prompt before the matching profanity.
The prompts are used as a prefix in the extraction of the profanity.
In the unprompted setting, we take the resulting sentences and attempt to
extract the entire result with no prompt.
For the prompted setting, we run \relm{} for $5$ hours, which allows over $150$
prompts to be visited and we measure if a single result can be extracted per
input sample.
These results are displayed in Figure~\ref{fig:relm_prompted}.
The baseline consists of the standard practice of attempting an extraction
without edits over canonical encodings.
We compare the baseline to \relm{}'s edit-distance preprocessor with Levenshtein distance 1 (\S\ref{sec:relm_API}),
which gives an additional degree of search freedom, in addition to enabling all encodings.
For the unprompted setting, we run \relm{} on all $2807$ matches in $4$ hours.
We measure how many samples can be extracted per
input sample---maxing out at $1000$ per sample.
We similarly compare with encodings and edits, and we measure the
\textit{total} number of token sequences extracted, rather than \textit{if} a
single example was extracted, as we did in the prompted case.
The results are shown in Figure~\ref{fig:relm_offensive_edits}.

\subsubsection{Qualitative Evaluation}
For prompted attacks, the easiest content to extract is nearly uniquely defined
as an insult.
Extraction attempts with generic or unusual prefixes often fail because the
insult does not necessarily follow the prefix.
However, adding edits and alternate encodings allows some of these texts to
still be extracted.
Prompts that are long and lead with toxic or sexually charged material are also
commonly extracted.
Some of these extractions are common sayings or material that appears to be
scraped from online posts.
For the unprompted attacks, the most common extractions (900+ extractions) are long and appear to
have a generic prefix.
We observe that enabling edits and all encodings enables some prompts to cover the
first
character of the bad word via edits, enabling extraction of the subsequent subword
tokens.
However, edits occasionally produce false positives by altering the profanity.
A common pattern is to border the query with special characters (e.g.,
\texttt{>},
\texttt{(},
\texttt{[},  
\texttt{?})
or
include special characters (e.g., \texttt{*}, \texttt{@}, \texttt{\#}, \texttt{-}) or
phonetic misspellings in the bad words.
See the appendix for an extended analysis with examples.

\observation{\relm{} extracts toxic samples by deriving
templates from a dataset. Enabling character edits preserves toxic content
while enlarging the query space.}

\subsubsection{Quantitative Evaluation}
In Figure~\ref{fig:relm_prompted}, we can see that prompted toxic content is
$2.5\times$ more successful using all of \relm{}'s encodings as well as
edits.
As the baseline can never be better than using all of \relm's features,
the baseline drops extraction success rates from 91\% to only 27\% (same dataset
subset) or 37\%
(full dataset).
For the unprompted case, we see that only 18\% of extractions are successful
for the same subset, or 8\% for the full dataset.
Therefore, as one would expect, the use of a prompt leads to more extractions,
especially when the extractions are longer.
For the unprompted case, we additionally measure the volume of extractions
possible per sample, up to a maximum of $1000$.
On average, $2.4$ samples are extracted per input.

For unprompted extractions (Figure~\ref{fig:relm_offensive_edits}), we can see that the bulk of results
come from edits.
Specifically,
97\% of extractions are from edits and
67\% are non-canonical.
Conditioning on both edits and encodings, we observe that only
1.1\% of returned results have no edits and are canonical,
31.7\% are canonical but have edits,
1.8\% are not canonical and have no edits, and
65.4\% are not canonical and have edits.
The most common additions/removals include:
\texttt{-},
\texttt{*},
\texttt{.},
\texttt{,},
\texttt{!},
\texttt{'},
\texttt{[},
\texttt{\#},
\texttt{@}
and
\texttt{i},
\texttt{"},
\texttt{u},
\texttt{,},
\texttt{c},
\texttt{e},
\texttt{f},
\texttt{b},
\texttt{o},
respectively. %

\observation{In the prompted setting, enabling edits and alternative encodings
unlocks $2.5\times$ more extractions per sample.
Doing the same in the unprompted setting results in $93\times$ more
examples of toxic content extractions per input, indicating verbatim toxicity generation may
severely underestimate toxicity exposure, especially for longer queries.}

\subsection{Testing for Language Understanding}%
\label{subsec:language_micro}
To see if \relm{} can be used as a tool for inference,
we revisit one of the benchmarks
used in the original GPT-2 paper~\cite{GPT2}.
Specifically, we focus on the LAMBADA dataset~\cite{paperno-etal-2016-lambada},
which measures long-range reasoning by measuring how accurately a model can
predict the last word, given a long string of context.
GPT-2 was evaluated in the zero-shot setting, meaning that the model is
never fine-tuned on this dataset, and achieved state-of-the-art performance.
Tuning LLM inference for this dataset is regarded as tricky~\cite{GPT3}, which
may require coding many different implementations to find the optimal solution.
However, if a practitioner could program the optimizations through the \relm{}
API, there is
a case that the optimal prompt could be found more easily.

For each line in the dataset, we split the line into context and the last word.
Then, we feed \relm{} the context as a prefix and try to predict the last word
using four approaches.
Intuitively, each of the approaches forces the completion to be a word
\code{[a-zA-Z]+} with optional punctuation and varying additional constraints.
First, with context \code{\textbf{<X>}}, we query \relm{} with
\code{\textbf{<X>}([a-zA-Z]+)(\textbackslash.|!|\textbackslash?)?(")?}, which we refer to as
\textit{baseline}.
Note that we escape \token{.} and \token{?} as literals with
\token{\textbackslash}, and we use \code{\textbf{<X>}} as a prefix.
We then query \relm{} with \textit{baseline} but with only the words in the
context used, as mentioned in the paper: \code{\textbf{<X>}
(\textbf{<words>})(\textbackslash.|!|\textbackslash?)?(")?}, where
\code{\textbf{<words>}} is the
set of words in context \code{\textbf{<X>}} separated by \code{|}.
We refer to this method as \textit{words}.
Next, we query \relm{} with \textit{baseline} but with \EOSToken{} concatenated
at the end, which we refer to as
\textit{terminated}.
Finally, we query \relm{} with \textit{terminated} but apply filters
(\S\ref{sec:relm_API}) to stop words, defined by \texttt{nltk}~\cite{nltk},
which we refer to as \textit{no\_stop}.
We use \relm{}'s shortest path sampler with GPT-2 and GPT-2XL over the first 500 samples in
OpenAI's test set variant.

\subsubsection{Qualitative Evaluation}
Filling in a blank with any string does not necessarily correspond
to the language of words,
because a string can be matched by a subword (\S\ref{sec:background}).
Even if a token represents a word, the model may be using it to
complete a sequence with additional words, rendering it invalid.
For \textit{baseline}, we get completions like \texttt{I can make it there on my
own, \underline{but}} (Shane)
or
\texttt{took a slow drag on [the cigarette] without ever taking his eyes off of,
\underline{J}} (Joran).
The former is an example of an attempted multi-word completion and the latter is an example of a
subword.
Using \textit{words} changes the first prediction to \texttt{I} (incorrect) and the latter
to \texttt{Joran} (correct), leading to a 15 point improvement with the addition of structure.
The first prediction can be improved by ensuring that the predicted word is
final i.e., \textit{terminated}---the addition of \EOSToken{} changes the former to be
\texttt{thanks}, which is still incorrect but is a reasonable last word completion.

Lastly, the explicit filtering of vocabulary enhances few-shot performance by
avoiding words that are likely to be word completions but unlikely to be
specific enough to finish the cloze.
For example, pronouns or \texttt{it} or \texttt{that} are too generic to be
correct: \texttt{Someone is to blame for what happened to \underline{her}} is
replaced with \texttt{Vivienne}.
Such stop-word filtering is not necessary for generic language modeling, but, in
this case, it's a useful bias to add into the model given the type of task that is being
performed.

\subsubsection{Quantitative Evaluation}
The first two approaches use the most common predictions ``the'', ``a'', ``her'', and ``and'' approximately
$12\%$ of the time in sum.
Adding \EOSToken{} makes the most common words ``him'', ``her'', ``me'', ``it'',
which account for $7\%$ of returns.
Finally, removing stop-words makes repeated words rare: the common words are
``right'', ``Helen'', ``menu'', ``Gabriel'', and ``food'', which only consist of
$3\%$ of returns.
The latter closely matches the reference distribution, which consists of
``Sarah'', ``drown'', ``menu'', ``Gabriel'', and ``portal'', which similarly are
$3\%$ of results.
The accuracy results for GPT-2 XL, along with GPT-2, are in Table~\ref{table:zero_shot}.
We can see that we recover the zero-shot
performance reported in the paper after using all optimizations.
Note that we even exceed the $63.24$ accuracy reported in the paper---these can
be either due to
\begin{inlineenum}
  \item the first $500$ samples being easier,
  \item minor differences in problem representation, as there is no publicly available
  reference implementation, and
  \item more thorough decode and search space semantics.
\end{inlineenum}
Regardless, we can see that tuning the regular expression in local ways results
in between 10~\cite{GPT2} and 30 point differences in zero-shot performance.

\begin{table}[h]
  \centering
  \begin{tabular}{lllll}
    \toprule
    model & baseline & words & terminated & no\_stop \\
    \midrule
    GPT-2XL & 41.6\% & 56.6\% & 65\% & 71\% \\
    \midrule
    GPT-2 & 27\% & 43\% & 46.4\%  & 52.2\%  \\
    \bottomrule
  \end{tabular}%
  \caption{Zero-shot LAMBADA accuracy on 500 examples.}
  \label{table:zero_shot}
\end{table}

\observation{\relm{} can enhance zero-shot accuracy by up to 30 points with minimal user effort
and without complex heuristics by injecting task constraints into the query.
}
\section{Additional Related Work}%
\label{sec:related_work}%

\textbf{Formal Languages in NLP.}
The \texttt{OpenGrm} library compiles regular expressions and
context-sensitive rewrite rules into automata, which can then be used to build
an n-gram model~\cite{roark-etal-2012-opengrm}.
Extensions to pushdown automata have similarly been used~\cite{allauzen-etal-2014-pushdown}.
More recently, a query language, LMQL, was proposed in~\cite{lmql}, exposing
both declarative SQL-like constructs as well as imperative ones to write LLM
prompt programs.
\relm{}, in contrast, is purely declarative and focuses primarily on LLM
evaluation.

\textbf{Adversarial Attacks and Controlled Generation in NLP.}
Adversarial attacks have been used to construct inputs into NLP systems, which
fool them into generating incorrect or harmful
content~\cite{wallace-etal-2019-universal,morris-etal-2020-textattack}.
Controlling the inference behavior of NLP models has prompted works in
fine-tuning model behavior~\cite{prabhumoye-etal-2020-exploring}.
One related idea to \relm{} is the use of a trie in decoding to enforce a
constrained beam search~\cite{decao2021autoregressive}.
Frameworks offer some tools to customize the set of tokens allowed during inference, though it is difficult
to make them work consistently due to tokenization ambiguities~\cite{huggingface_bad_words}.
\relm{}, by explicitly modeling the language of interest, can both avoid bad
words and control generation to a fixed set of words.
The most related work to \relm{} that we are aware of is
CheckList~\cite{checklist}, which uses templates to test a language model.
\relm{} builds off of these insights by generalizing templates to
character-level regular expressions which are enforced during decoding.

\section{Conclusion}%
\label{sec:conclusion}%
The complexity of natural language combined with rapid progress in large
language models (LLMs) has resulted in a need for LLM validation abstractions.
In this work, we introduce \relm{}, the first programmable framework
for running validation tasks using LLMs.
\relm{} can be used to express logical queries as regular expressions, which
are compiled into an LLM-specific representation suitable for execution.
Over memorization, gender bias, toxicity, and language understanding tasks,
\relm{} is able to
execute queries up to $15\times$ faster, with $2.5\times$ less data, or in a manner that enables 
additional insights.
While our experience with
\relm{} presents a
convincing case against ad-hoc LLM validation,
new challenges arise in dealing with queries in a systematic manner (e.g.,
left-to-right autoregressive decoding has an affinity toward
suffix completions).
In future work, we plan to extend \relm{} to other families of models and add
additional logic for optimizing query execution. %
 
\section*{Acknowledgements}
We thank the anonymous reviewers for their help improving the presentation of this paper.
We also thank the members and companies of the PDL Consortium: Amazon, Google, Hewlett Packard Enterprise, Hitachi Ltd., IBM Research, Intel Corporation, Meta, Microsoft Research, NetApp, Inc., Oracle Corporation, Pure Storage, Salesforce, Samsung Semiconductor Inc., Seagate Technology, Two Sigma, and Western Digital for their interest, insights, feedback, and support. 

This material is based upon work supported by the U.S. Army Research Office and the U.S. Army Futures Command under Contract No. W911NF-20-D-0002.
The content of the information does not necessarily reflect the position or the policy of the government and no official endorsement should be inferred. 

\clearpage

\bibliography{ms}

\begin{thebibliography}{66}
\providecommand{\natexlab}[1]{#1}
\providecommand{\url}[1]{\texttt{#1}}
\expandafter\ifx\csname urlstyle\endcsname\relax
  \providecommand{\doi}[1]{doi: #1}\else
  \providecommand{\doi}{doi: \begingroup \urlstyle{rm}\Url}\fi

\bibitem[Allauzen et~al.(2007)Allauzen, Riley, Schalkwyk, Skut, and
  Mohri]{allauzen2007openfst}
Allauzen, C., Riley, M., Schalkwyk, J., Skut, W., and Mohri, M.
\newblock Openfst: A general and efficient weighted finite-state transducer
  library.
\newblock In \emph{International Conference on Implementation and Application
  of Automata}, pp.\  11--23. Springer, 2007.

\bibitem[Allauzen et~al.(2014)Allauzen, Byrne, de~Gispert, Iglesias, and
  Riley]{allauzen-etal-2014-pushdown}
Allauzen, C., Byrne, B., de~Gispert, A., Iglesias, G., and Riley, M.
\newblock Pushdown automata in statistical machine translation.
\newblock \emph{Computational Linguistics}, 40\penalty0 (3):\penalty0 687--723,
  September 2014.
\newblock \doi{10.1162/COLI_a_00197}.

\bibitem[Bender et~al.(2021)Bender, Gebru, McMillan-Major, and
  Shmitchell]{bender2021dangers}
Bender, E.~M., Gebru, T., McMillan-Major, A., and Shmitchell, S.
\newblock On the dangers of stochastic parrots: Can language models be too big?
\newblock In \emph{Proceedings of the ACM Conference on Fairness,
  Accountability, and Transparency}, pp.\  610--623, 2021.

\bibitem[Beurer-Kellner et~al.(2022)Beurer-Kellner, Fischer, and Vechev]{lmql}
Beurer-Kellner, L., Fischer, M., and Vechev, M.
\newblock Prompting is programming: A query language for large language models.
\newblock \emph{arXiv preprint arXiv:2212.06094}, 2022.

\bibitem[Beutel et~al.(2020)Beutel, Chi, Pavlick, Pitler, Tenney, Chen,
  Webster, Petrov, and Wang]{measuring_gendered_correlations}
Beutel, A., Chi, E.~H., Pavlick, E., Pitler, E.~B., Tenney, I., Chen, J.,
  Webster, K., Petrov, S., and Wang, X.
\newblock Measuring and reducing gendered correlations in pre-trained models.
\newblock \emph{arXiv preprint arXiv:2010.06032}, 2020.

\bibitem[Bird et~al.(2009)Bird, Loper, and Klein]{nltk}
Bird, S., Loper, E., and Klein, E.
\newblock \emph{Natural Language Processing with Python}.
\newblock O'Reilly Media Inc., 2009.

\bibitem[Bommasani et~al.(2021)Bommasani, Hudson, Adeli, Altman, Arora, von
  Arx, Bernstein, Bohg, Bosselut, Brunskill, Brynjolfsson, Buch, Card,
  Castellon, Chatterji, Chen, Creel, Davis, Demszky, Donahue, Doumbouya,
  Durmus, Ermon, Etchemendy, Ethayarajh, Fei-Fei, Finn, Gale, Gillespie, Goel,
  Goodman, Grossman, Guha, Hashimoto, Henderson, Hewitt, Ho, Hong, Hsu, Huang,
  Icard, Jain, Jurafsky, Kalluri, Karamcheti, Keeling, Khani, Khattab, Koh,
  Krass, Krishna, Kuditipudi, Kumar, Ladhak, Lee, Lee, Leskovec, Levent, Li,
  Li, Ma, Malik, Manning, Mirchandani, Mitchell, Munyikwa, Nair, Narayan,
  Narayanan, Newman, Nie, Niebles, Nilforoshan, Nyarko, Ogut, Orr,
  Papadimitriou, Park, Piech, Portelance, Potts, Raghunathan, Reich, Ren, Rong,
  Roohani, Ruiz, Ryan, Ré, Sadigh, Sagawa, Santhanam, Shih, Srinivasan,
  Tamkin, Taori, Thomas, Tramèr, Wang, Wang, Wu, Wu, Wu, Xie, Yasunaga, You,
  Zaharia, Zhang, Zhang, Zhang, Zhang, Zheng, Zhou, and
  Liang]{bommasani2021opportunities}
Bommasani, R., Hudson, D.~A., Adeli, E., Altman, R., Arora, S., von Arx, S.,
  Bernstein, M.~S., Bohg, J., Bosselut, A., Brunskill, E., Brynjolfsson, E.,
  Buch, S., Card, D., Castellon, R., Chatterji, N., Chen, A., Creel, K., Davis,
  J.~Q., Demszky, D., Donahue, C., Doumbouya, M., Durmus, E., Ermon, S.,
  Etchemendy, J., Ethayarajh, K., Fei-Fei, L., Finn, C., Gale, T., Gillespie,
  L., Goel, K., Goodman, N., Grossman, S., Guha, N., Hashimoto, T., Henderson,
  P., Hewitt, J., Ho, D.~E., Hong, J., Hsu, K., Huang, J., Icard, T., Jain, S.,
  Jurafsky, D., Kalluri, P., Karamcheti, S., Keeling, G., Khani, F., Khattab,
  O., Koh, P.~W., Krass, M., Krishna, R., Kuditipudi, R., Kumar, A., Ladhak,
  F., Lee, M., Lee, T., Leskovec, J., Levent, I., Li, X.~L., Li, X., Ma, T.,
  Malik, A., Manning, C.~D., Mirchandani, S., Mitchell, E., Munyikwa, Z., Nair,
  S., Narayan, A., Narayanan, D., Newman, B., Nie, A., Niebles, J.~C.,
  Nilforoshan, H., Nyarko, J., Ogut, G., Orr, L., Papadimitriou, I., Park,
  J.~S., Piech, C., Portelance, E., Potts, C., Raghunathan, A., Reich, R., Ren,
  H., Rong, F., Roohani, Y., Ruiz, C., Ryan, J., Ré, C., Sadigh, D., Sagawa,
  S., Santhanam, K., Shih, A., Srinivasan, K., Tamkin, A., Taori, R., Thomas,
  A.~W., Tramèr, F., Wang, R.~E., Wang, W., Wu, B., Wu, J., Wu, Y., Xie,
  S.~M., Yasunaga, M., You, J., Zaharia, M., Zhang, M., Zhang, T., Zhang, X.,
  Zhang, Y., Zheng, L., Zhou, K., and Liang, P.
\newblock On the opportunities and risks of foundation models.
\newblock \emph{arXiv preprint arXiv:2108.07258}, 2021.

\bibitem[Bowman \& Dahl(2021)Bowman and Dahl]{what_will_it_take_fix_nlp}
Bowman, S.~R. and Dahl, G.~E.
\newblock What will it take to fix benchmarking in natural language
  understanding?
\newblock In \emph{Proceedings of the 2021 Conference of the North American
  Chapter of the Association for Computational Linguistics: Human Language
  Technologies}, pp.\  4843--4855. Association for Computational Linguistics,
  2021.

\bibitem[Brown et~al.(2020)Brown, Mann, Ryder, Subbiah, Kaplan, Dhariwal,
  Neelakantan, Shyam, Sastry, Askell, Agarwal, Herbert-Voss, Krueger, Henighan,
  Child, Ramesh, Ziegler, Wu, Winter, Hesse, Chen, Sigler, Litwin, Gray, Chess,
  Clark, Berner, McCandlish, Radford, Sutskever, and Amodei]{GPT3}
Brown, T., Mann, B., Ryder, N., Subbiah, M., Kaplan, J.~D., Dhariwal, P.,
  Neelakantan, A., Shyam, P., Sastry, G., Askell, A., Agarwal, S.,
  Herbert-Voss, A., Krueger, G., Henighan, T., Child, R., Ramesh, A., Ziegler,
  D., Wu, J., Winter, C., Hesse, C., Chen, M., Sigler, E., Litwin, M., Gray,
  S., Chess, B., Clark, J., Berner, C., McCandlish, S., Radford, A., Sutskever,
  I., and Amodei, D.
\newblock Language models are few-shot learners.
\newblock In \emph{Advances in Neural Information Processing Systems}, 2020.

\bibitem[Carlini et~al.(2019)Carlini, Liu, Erlingsson, Kos, and
  Song]{secret_sharer}
Carlini, N., Liu, C., Erlingsson, U., Kos, J., and Song, D.
\newblock The secret sharer: Evaluating and testing unintended memorization in
  neural networks.
\newblock In \emph{USENIX Security Symposium}, pp.\  267--284, 2019.

\bibitem[Carlini et~al.(2021)Carlini, Tramer, Wallace, Jagielski, Herbert-Voss,
  Lee, Roberts, Brown, Song, Erlingsson, Oprea, and
  Raffel]{carlini2021extracting}
Carlini, N., Tramer, F., Wallace, E., Jagielski, M., Herbert-Voss, A., Lee, K.,
  Roberts, A., Brown, T., Song, D., Erlingsson, U., Oprea, A., and Raffel, C.
\newblock Extracting training data from large language models.
\newblock In \emph{USENIX Security Symposium}, pp.\  2633--2650, 2021.

\bibitem[Carlini et~al.(2023)Carlini, Ippolito, Jagielski, Lee, Tramer, and
  Zhang]{carlini2022quantifying}
Carlini, N., Ippolito, D., Jagielski, M., Lee, K., Tramer, F., and Zhang, C.
\newblock Quantifying memorization across neural language models.
\newblock In \emph{International Conference on Learning Representations}, 2023.

\bibitem[Chouldechova \& Roth(2020)Chouldechova and
  Roth]{chouldechova2020snapshot}
Chouldechova, A. and Roth, A.
\newblock A snapshot of the frontiers of fairness in machine learning.
\newblock \emph{Communications of the ACM}, 63\penalty0 (5):\penalty0 82--89,
  2020.

\bibitem[Chowdhery et~al.(2022)Chowdhery, Narang, Devlin, Bosma, Mishra,
  Roberts, Barham, Chung, Sutton, Gehrmann, Schuh, Shi, Tsvyashchenko, Maynez,
  Rao, Barnes, Tay, Shazeer, Prabhakaran, Reif, Du, Hutchinson, Pope, Bradbury,
  Austin, Isard, Gur-Ari, Yin, Duke, Levskaya, Ghemawat, Dev, Michalewski,
  Garcia, Misra, Robinson, Fedus, Zhou, Ippolito, Luan, Lim, Zoph, Spiridonov,
  Sepassi, Dohan, Agrawal, Omernick, Dai, Pillai, Pellat, Lewkowycz, Moreira,
  Child, Polozov, Lee, Zhou, Wang, Saeta, Diaz, Firat, Catasta, Wei,
  Meier-Hellstern, Eck, Dean, Petrov, and Fiedel]{PALM}
Chowdhery, A., Narang, S., Devlin, J., Bosma, M., Mishra, G., Roberts, A.,
  Barham, P., Chung, H.~W., Sutton, C., Gehrmann, S., Schuh, P., Shi, K.,
  Tsvyashchenko, S., Maynez, J., Rao, A., Barnes, P., Tay, Y., Shazeer, N.,
  Prabhakaran, V., Reif, E., Du, N., Hutchinson, B., Pope, R., Bradbury, J.,
  Austin, J., Isard, M., Gur-Ari, G., Yin, P., Duke, T., Levskaya, A.,
  Ghemawat, S., Dev, S., Michalewski, H., Garcia, X., Misra, V., Robinson, K.,
  Fedus, L., Zhou, D., Ippolito, D., Luan, D., Lim, H., Zoph, B., Spiridonov,
  A., Sepassi, R., Dohan, D., Agrawal, S., Omernick, M., Dai, A.~M., Pillai,
  T.~S., Pellat, M., Lewkowycz, A., Moreira, E., Child, R., Polozov, O., Lee,
  K., Zhou, Z., Wang, X., Saeta, B., Diaz, M., Firat, O., Catasta, M., Wei, J.,
  Meier-Hellstern, K., Eck, D., Dean, J., Petrov, S., and Fiedel, N.
\newblock Palm: Scaling language modeling with pathways.
\newblock \emph{arXiv preprint arXiv:2204.02311}, 2022.

\bibitem[{De Cao} et~al.(2021){De Cao}, Izacard, Riedel, and
  Petroni]{decao2021autoregressive}
{De Cao}, N., Izacard, G., Riedel, S., and Petroni, F.
\newblock Autoregressive entity retrieval.
\newblock In \emph{International Conference on Learning Representations}, 2021.

\bibitem[Devlin et~al.(2019)Devlin, Chang, Lee, and Toutanova]{BERT}
Devlin, J., Chang, M., Lee, K., and Toutanova, K.
\newblock {BERT:} pre-training of deep bidirectional transformers for language
  understanding.
\newblock In \emph{Proceedings of the 2019 Conference of the North American
  Chapter of the Association for Computational Linguistics: Human Language
  Technologies}, pp.\  4171--4186. Association for Computational Linguistics,
  2019.

\bibitem[Dijkstra(2022)]{dijkstras_shortest_path}
Dijkstra, E.~W.
\newblock A note on two problems in connexion with graphs.
\newblock In \emph{Edsger Wybe Dijkstra: His Life, Work, and Legacy}, pp.\
  287--290. 2022.

\bibitem[Dunietz et~al.(2020)Dunietz, Burnham, Bharadwaj, Rambow, Chu-Carroll,
  and Ferrucci]{dunietz-etal-2020-test}
Dunietz, J., Burnham, G., Bharadwaj, A., Rambow, O., Chu-Carroll, J., and
  Ferrucci, D.
\newblock To test machine comprehension, start by defining comprehension.
\newblock In \emph{Proceedings of the 58th Annual Meeting of the Association
  for Computational Linguistics}, pp.\  7839--7859, Online, July 2020.
  Association for Computational Linguistics.
\newblock \doi{10.18653/v1/2020.acl-main.701}.

\bibitem[Fan et~al.(2018)Fan, Lewis, and Dauphin]{fan-etal-2018-hierarchical}
Fan, A., Lewis, M., and Dauphin, Y.
\newblock Hierarchical neural story generation.
\newblock In \emph{Proceedings of the 56th Annual Meeting of the Association
  for Computational Linguistics (Volume 1: Long Papers)}, pp.\  889--898.
  Association for Computational Linguistics, 2018.

\bibitem[Gage(1994)]{BPE}
Gage, P.
\newblock A new algorithm for data compression.
\newblock
  \url{https://www.drdobbs.com/a-new-algorithm-for-data-compression/184402829},
  1994.

\bibitem[Gao et~al.(2020)Gao, Biderman, Black, Golding, Hoppe, Foster, Phang,
  He, Thite, Nabeshima, Presser, and Leahy]{pile}
Gao, L., Biderman, S., Black, S., Golding, L., Hoppe, T., Foster, C., Phang,
  J., He, H., Thite, A., Nabeshima, N., Presser, S., and Leahy, C.
\newblock The {P}ile: An 800gb dataset of diverse text for language modeling.
\newblock \emph{arXiv preprint arXiv:2101.00027}, 2020.

\bibitem[Gao et~al.(2021{\natexlab{a}})Gao, Tow, Biderman, Black, DiPofi,
  Foster, Golding, Hsu, McDonell, Muennighoff, Phang, Reynolds, Tang, Thite,
  Wang, Wang, and Zou]{eval-harness}
Gao, L., Tow, J., Biderman, S., Black, S., DiPofi, A., Foster, C., Golding, L.,
  Hsu, J., McDonell, K., Muennighoff, N., Phang, J., Reynolds, L., Tang, E.,
  Thite, A., Wang, B., Wang, K., and Zou, A.
\newblock A framework for few-shot language model evaluation.
\newblock \url{https://github.com/EleutherAI/lm-evaluation-harness}, September
  2021{\natexlab{a}}.

\bibitem[Gao et~al.(2021{\natexlab{b}})Gao, Fisch, and
  Chen]{gao-etal-2021-making}
Gao, T., Fisch, A., and Chen, D.
\newblock Making pre-trained language models better few-shot learners.
\newblock In \emph{Proceedings of the 59th Annual Meeting of the Association
  for Computational Linguistics and the 11th International Joint Conference on
  Natural Language Processing (Volume 1: Long Papers)}, pp.\  3816--3830,
  Online, August 2021{\natexlab{b}}. Association for Computational Linguistics.
\newblock \doi{10.18653/v1/2021.acl-long.295}.

\bibitem[Gehman et~al.(2020)Gehman, Gururangan, Sap, Choi, and
  Smith]{gehman-etal-2020-realtoxicityprompts}
Gehman, S., Gururangan, S., Sap, M., Choi, Y., and Smith, N.~A.
\newblock {R}eal{T}oxicity{P}rompts: Evaluating neural toxic degeneration in
  language models.
\newblock In \emph{Findings of the Association for Computational Linguistics:
  EMNLP 2020}, pp.\  3356--3369. Association for Computational Linguistics,
  2020.

\bibitem[Gorman(2016)]{gorman-2016-pynini}
Gorman, K.
\newblock {P}ynini: A {P}ython library for weighted finite-state grammar
  compilation.
\newblock In \emph{Proceedings of the {SIGFSM} Workshop on Statistical {NLP}
  and Weighted Automata}, pp.\  75--80, Berlin, Germany, August 2016.
  Association for Computational Linguistics.
\newblock \doi{10.18653/v1/W16-2409}.
\newblock URL \url{https://aclanthology.org/W16-2409}.

\bibitem[Hartvigsen et~al.(2022)Hartvigsen, Gabriel, Palangi, Sap, Ray, and
  Kamar]{hartvigsen-etal-2022-toxigen}
Hartvigsen, T., Gabriel, S., Palangi, H., Sap, M., Ray, D., and Kamar, E.
\newblock {T}oxi{G}en: A large-scale machine-generated dataset for adversarial
  and implicit hate speech detection.
\newblock In \emph{Proceedings of the 60th Annual Meeting of the Association
  for Computational Linguistics (Volume 1: Long Papers)}, pp.\  3309--3326.
  Association for Computational Linguistics, 2022.

\bibitem[Hassan et~al.(2008)Hassan, Noeman, and Hassan]{hassan2008language}
Hassan, A., Noeman, S., and Hassan, H.
\newblock Language independent text correction using finite state automata.
\newblock In \emph{Proceedings of the Third International Joint Conference on
  Natural Language Processing: Volume-II}, 2008.

\bibitem[Holtzman et~al.(2020)Holtzman, Buys, Du, Forbes, and
  Choi]{Holtzman2020The}
Holtzman, A., Buys, J., Du, L., Forbes, M., and Choi, Y.
\newblock The curious case of neural text degeneration.
\newblock In \emph{International Conference on Learning Representations}, 2020.

\bibitem[Hopcroft et~al.(2007)Hopcroft, Motwani, and Ullman]{automatalanguages}
Hopcroft, J.~E., Motwani, R., and Ullman, J.~D.
\newblock \emph{Introduction to Automata Theory, Languages, and Computation}.
\newblock 3rd edition, 2007.

\bibitem[HuggingFace(2021)]{huggingfacegeneration}
HuggingFace.
\newblock run\_generation.py.
\newblock
  \url{https://github.com/huggingface/transformers/blob/main/examples/pytorch/text-generation/run_generation.py},
  2021.

\bibitem[{Huggingface}(2022)]{huggingface_bad_words}
{Huggingface}.
\newblock {bad\_words\_ids not working}.
\newblock \url{https://github.com/huggingface/transformers/issues/17504}, 2022.

\bibitem[Jiang et~al.(2020)Jiang, Xu, Araki, and Neubig]{jiang-etal-2020-know}
Jiang, Z., Xu, F.~F., Araki, J., and Neubig, G.
\newblock How can we know what language models know?
\newblock \emph{Transactions of the Association for Computational Linguistics},
  8:\penalty0 423--438, 2020.
\newblock \doi{10.1162/tacl_a_00324}.

\bibitem[Kiela et~al.(2021)Kiela, Bartolo, Nie, Kaushik, Geiger, Wu, Vidgen,
  Prasad, Singh, Ringshia, Ma, Thrush, Riedel, Waseem, Stenetorp, Jia, Bansal,
  Potts, and Williams]{kiela-etal-2021-dynabench}
Kiela, D., Bartolo, M., Nie, Y., Kaushik, D., Geiger, A., Wu, Z., Vidgen, B.,
  Prasad, G., Singh, A., Ringshia, P., Ma, Z., Thrush, T., Riedel, S., Waseem,
  Z., Stenetorp, P., Jia, R., Bansal, M., Potts, C., and Williams, A.
\newblock Dynabench: Rethinking benchmarking in {NLP}.
\newblock In \emph{Proceedings of the 2021 Conference of the North American
  Chapter of the Association for Computational Linguistics: Human Language
  Technologies}, pp.\  4110--4124. Association for Computational Linguistics,
  2021.

\bibitem[Kirk et~al.(2021)Kirk, Jun, Volpin, Iqbal, Benussi, Dreyer,
  Shtedritski, and Asano]{bias_out_of_box}
Kirk, H.~R., Jun, Y., Volpin, F., Iqbal, H., Benussi, E., Dreyer, F.,
  Shtedritski, A., and Asano, Y.
\newblock Bias out-of-the-box: An empirical analysis of intersectional
  occupational biases in popular generative language models.
\newblock In \emph{Advances in Neural Information Processing Systems}, 2021.

\bibitem[Kumar et~al.(2021)Kumar, Kelley, Consolvo, Mason, Bursztein,
  Durumeric, Thomas, and Bailey]{designing_toxic_content_classification}
Kumar, D., Kelley, P.~G., Consolvo, S., Mason, J., Bursztein, E., Durumeric,
  Z., Thomas, K., and Bailey, M.
\newblock Designing toxic content classification for a diversity of
  perspectives.
\newblock In \emph{Seventeenth Symposium on Usable Privacy and Security (SOUPS
  2021)}, pp.\  299--318. USENIX Association, August 2021.
\newblock ISBN 978-1-939133-25-0.

\bibitem[Kurita et~al.(2019)Kurita, Vyas, Pareek, Black, and
  Tsvetkov]{kurita-etal-2019-measuring}
Kurita, K., Vyas, N., Pareek, A., Black, A.~W., and Tsvetkov, Y.
\newblock Measuring bias in contextualized word representations.
\newblock In \emph{Proceedings of the First Workshop on Gender Bias in Natural
  Language Processing}, pp.\  166--172. Association for Computational
  Linguistics, 2019.

\bibitem[Liang et~al.(2022)Liang, Bommasani, Lee, Tsipras, Soylu, Yasunaga,
  Zhang, Narayanan, Wu, Kumar, Newman, Yuan, Yan, Zhang, Cosgrove, Manning,
  Ré, Acosta-Navas, Hudson, Zelikman, Durmus, Ladhak, Rong, Ren, Yao, Wang,
  Santhanam, Orr, Zheng, Yuksekgonul, Suzgun, Kim, Guha, Chatterji, Khattab,
  Henderson, Huang, Chi, Xie, Santurkar, Ganguli, Hashimoto, Icard, Zhang,
  Chaudhary, Wang, Li, Mai, Zhang, and Koreeda]{helm}
Liang, P., Bommasani, R., Lee, T., Tsipras, D., Soylu, D., Yasunaga, M., Zhang,
  Y., Narayanan, D., Wu, Y., Kumar, A., Newman, B., Yuan, B., Yan, B., Zhang,
  C., Cosgrove, C., Manning, C.~D., Ré, C., Acosta-Navas, D., Hudson, D.~A.,
  Zelikman, E., Durmus, E., Ladhak, F., Rong, F., Ren, H., Yao, H., Wang, J.,
  Santhanam, K., Orr, L., Zheng, L., Yuksekgonul, M., Suzgun, M., Kim, N.,
  Guha, N., Chatterji, N., Khattab, O., Henderson, P., Huang, Q., Chi, R., Xie,
  S.~M., Santurkar, S., Ganguli, S., Hashimoto, T., Icard, T., Zhang, T.,
  Chaudhary, V., Wang, W., Li, X., Mai, Y., Zhang, Y., and Koreeda, Y.
\newblock Holistic evaluation of language models.
\newblock \emph{arXiv preprint arXiv:2211.09110}, 2022.

\bibitem[Liu et~al.(2021)Liu, Zheng, Du, Ding, Qian, Yang, and
  Tang]{liu2021gpt}
Liu, X., Zheng, Y., Du, Z., Ding, M., Qian, Y., Yang, Z., and Tang, J.
\newblock Gpt understands, too.
\newblock \emph{arXiv preprint arXiv:2103.10385}, 2021.

\bibitem[Mihov \& Schulz(2019)Mihov and Schulz]{mihov_schulz_2019}
Mihov, S. and Schulz, K.~U.
\newblock \emph{Finite-State Techniques: Automata, Transducers and Bimachines}.
\newblock Cambridge Tracts in Theoretical Computer Science. Cambridge
  University Press, 2019.
\newblock \doi{10.1017/9781108756945}.

\bibitem[Mohri(1997)]{mohri-1997-finite}
Mohri, M.
\newblock Finite-state transducers in language and speech processing.
\newblock \emph{Computational Linguistics}, 23\penalty0 (2):\penalty0 269--311,
  1997.

\bibitem[Morris et~al.(2020)Morris, Lifland, Yoo, Grigsby, Jin, and
  Qi]{morris-etal-2020-textattack}
Morris, J., Lifland, E., Yoo, J.~Y., Grigsby, J., Jin, D., and Qi, Y.
\newblock {T}ext{A}ttack: A framework for adversarial attacks, data
  augmentation, and adversarial training in {NLP}.
\newblock In \emph{Proceedings of the 2020 Conference on Empirical Methods in
  Natural Language Processing: System Demonstrations}, pp.\  119--126, Online,
  October 2020. Association for Computational Linguistics.
\newblock \doi{10.18653/v1/2020.emnlp-demos.16}.

\bibitem[Ousidhoum et~al.(2021)Ousidhoum, Zhao, Fang, Song, and
  Yeung]{ousidhoum-etal-2021-probing}
Ousidhoum, N., Zhao, X., Fang, T., Song, Y., and Yeung, D.-Y.
\newblock Probing toxic content in large pre-trained language models.
\newblock In \emph{Proceedings of the 59th Annual Meeting of the Association
  for Computational Linguistics and the 11th International Joint Conference on
  Natural Language Processing (Volume 1: Long Papers)}, pp.\  4262--4274,
  Online, August 2021. Association for Computational Linguistics.
\newblock \doi{10.18653/v1/2021.acl-long.329}.

\bibitem[Paperno et~al.(2016)Paperno, Kruszewski, Lazaridou, Pham, Bernardi,
  Pezzelle, Baroni, Boleda, and Fern{\'a}ndez]{paperno-etal-2016-lambada}
Paperno, D., Kruszewski, G., Lazaridou, A., Pham, N.~Q., Bernardi, R.,
  Pezzelle, S., Baroni, M., Boleda, G., and Fern{\'a}ndez, R.
\newblock The {LAMBADA} dataset: Word prediction requiring a broad discourse
  context.
\newblock In \emph{Proceedings of the 54th Annual Meeting of the Association
  for Computational Linguistics (Volume 1: Long Papers)}, pp.\  1525--1534,
  Berlin, Germany, August 2016. Association for Computational Linguistics.
\newblock \doi{10.18653/v1/P16-1144}.

\bibitem[Paszke et~al.(2019)Paszke, Gross, Massa, Lerer, Bradbury, Chanan,
  Killeen, Lin, Gimelshein, Antiga, et~al.]{paszke2019pytorch}
Paszke, A., Gross, S., Massa, F., Lerer, A., Bradbury, J., Chanan, G., Killeen,
  T., Lin, Z., Gimelshein, N., Antiga, L., et~al.
\newblock Pytorch: An imperative style, high-performance deep learning library.
\newblock 2019.

\bibitem[Pereira \& Riley(1996)Pereira and Riley]{SpeechRecognitionComposition}
Pereira, F. C.~N. and Riley, M.~D.
\newblock Speech recognition by composition of weighted finite automata.
\newblock In \emph{Finite-State Language Processing}, pp.\  431--453, 1996.

\bibitem[Prabhumoye et~al.(2020)Prabhumoye, Black, and
  Salakhutdinov]{prabhumoye-etal-2020-exploring}
Prabhumoye, S., Black, A.~W., and Salakhutdinov, R.
\newblock Exploring controllable text generation techniques.
\newblock In \emph{Proceedings of the 28th International Conference on
  Computational Linguistics}, pp.\  1--14, Barcelona, Spain (Online), December
  2020. International Committee on Computational Linguistics.
\newblock \doi{10.18653/v1/2020.coling-main.1}.

\bibitem[Prager(2006)]{Prager2006OpenDomainQ}
Prager, J.~M.
\newblock Open-domain question-answering.
\newblock \emph{Found. Trends Inf. Retr.}, 1:\penalty0 91--231, 2006.

\bibitem[Radford et~al.(2018)Radford, Narasimhan, Salimans, and
  Sutskever]{GPT1}
Radford, A., Narasimhan, K., Salimans, T., and Sutskever, I.
\newblock Improving language understanding by generative pre-training.
\newblock 2018.

\bibitem[Radford et~al.(2019)Radford, Wu, Child, Luan, Amodei, and
  Sutskever]{GPT2}
Radford, A., Wu, J., Child, R., Luan, D., Amodei, D., and Sutskever, I.
\newblock Language models are unsupervised multitask learners.
\newblock 2019.

\bibitem[Raffel et~al.(2020)Raffel, Shazeer, Roberts, Lee, Narang, Matena,
  Zhou, Li, and Liu]{JMLR:T5}
Raffel, C., Shazeer, N., Roberts, A., Lee, K., Narang, S., Matena, M., Zhou,
  Y., Li, W., and Liu, P.~J.
\newblock Exploring the limits of transfer learning with a unified text-to-text
  transformer.
\newblock \emph{Journal of Machine Learning Research}, 21\penalty0
  (140):\penalty0 1--67, 2020.

\bibitem[Reynolds \& McDonell(2021)Reynolds and McDonell]{promptprogramming}
Reynolds, L. and McDonell, K.
\newblock Prompt programming for large language models: Beyond the few-shot
  paradigm.
\newblock In \emph{Extended Abstracts of the 2021 CHI Conference on Human
  Factors in Computing Systems}, CHI EA '21, New York, NY, USA, 2021.
  Association for Computing Machinery.
\newblock ISBN 9781450380959.
\newblock \doi{10.1145/3411763.3451760}.

\bibitem[Ribeiro et~al.(2020)Ribeiro, Wu, Guestrin, and Singh]{checklist}
Ribeiro, M.~T., Wu, T., Guestrin, C., and Singh, S.
\newblock Beyond accuracy: Behavioral testing of {NLP} models with
  {C}heck{L}ist.
\newblock In \emph{Proceedings of the 58th Annual Meeting of the Association
  for Computational Linguistics}, pp.\  4902--4912, Online, July 2020.
  Association for Computational Linguistics.
\newblock \doi{10.18653/v1/2020.acl-main.442}.
\newblock URL \url{https://aclanthology.org/2020.acl-main.442}.

\bibitem[Roark et~al.(2012)Roark, Sproat, Allauzen, Riley, Sorensen, and
  Tai]{roark-etal-2012-opengrm}
Roark, B., Sproat, R., Allauzen, C., Riley, M., Sorensen, J., and Tai, T.
\newblock The {O}pen{G}rm open-source finite-state grammar software libraries.
\newblock In \emph{Proceedings of the {ACL} 2012 System Demonstrations}, pp.\
  61--66, Jeju Island, Korea, July 2012. Association for Computational
  Linguistics.

\bibitem[Roberts et~al.(2020)Roberts, Raffel, and
  Shazeer]{roberts-etal-2020-much}
Roberts, A., Raffel, C., and Shazeer, N.
\newblock How much knowledge can you pack into the parameters of a language
  model?
\newblock In \emph{Proceedings of the 2020 Conference on Empirical Methods in
  Natural Language Processing (EMNLP)}, pp.\  5418--5426, Online, November
  2020. Association for Computational Linguistics.
\newblock \doi{10.18653/v1/2020.emnlp-main.437}.
\newblock URL \url{https://aclanthology.org/2020.emnlp-main.437}.

\bibitem[Rockt{\"a}schel et~al.(2016)Rockt{\"a}schel, Grefenstette, Hermann,
  Ko{\v{c}}isk{\`y}, and Blunsom]{rocktaschel2015reasoning}
Rockt{\"a}schel, T., Grefenstette, E., Hermann, K.~M., Ko{\v{c}}isk{\`y}, T.,
  and Blunsom, P.
\newblock Reasoning about entailment with neural attention.
\newblock In \emph{International Conference on Learning Representations}, 2016.

\bibitem[Schick \& Sch{\"u}tze(2021)Schick and
  Sch{\"u}tze]{schick-schutze-2021-exploiting}
Schick, T. and Sch{\"u}tze, H.
\newblock Exploiting cloze-questions for few-shot text classification and
  natural language inference.
\newblock In \emph{Proceedings of the 16th Conference of the European Chapter
  of the Association for Computational Linguistics: Main Volume}, pp.\
  255--269, Online, April 2021. Association for Computational Linguistics.
\newblock \doi{10.18653/v1/2021.eacl-main.20}.

\bibitem[Schuurmans(2023)]{schuurmans2023memory}
Schuurmans, D.
\newblock Memory augmented large language models are computationally universal.
\newblock \emph{arXiv preprint arXiv:2301.04589}, 2023.

\bibitem[Sheng et~al.(2019)Sheng, Chang, Natarajan, and
  Peng]{thewomanworkedasbabysitter}
Sheng, E., Chang, K.-W., Natarajan, P., and Peng, N.
\newblock The woman worked as a babysitter: On biases in language generation.
\newblock In \emph{Proceedings of the 2019 Conference on Empirical Methods in
  Natural Language Processing and the 9th International Joint Conference on
  Natural Language Processing (EMNLP-IJCNLP)}, pp.\  3407--3412, Hong Kong,
  China, November 2019. Association for Computational Linguistics.
\newblock \doi{10.18653/v1/D19-1339}.
\newblock URL \url{https://aclanthology.org/D19-1339}.

\bibitem[Srivastava et~al.(2022)Srivastava, Rastogi, Rao, Shoeb, Abid, Fisch,
  Brown, Santoro, Gupta, Garriga-Alonso, et~al.]{srivastava2022beyond}
Srivastava, A., Rastogi, A., Rao, A., Shoeb, A. A.~M., Abid, A., Fisch, A.,
  Brown, A.~R., Santoro, A., Gupta, A., Garriga-Alonso, A., et~al.
\newblock Beyond the imitation game: Quantifying and extrapolating the
  capabilities of language models.
\newblock \emph{arXiv preprint arXiv:2206.04615}, 2022.

\bibitem[Sugawara et~al.(2020)Sugawara, Stenetorp, Inui, and
  Aizawa]{sugawara2020assessing}
Sugawara, S., Stenetorp, P., Inui, K., and Aizawa, A.
\newblock Assessing the benchmarking capacity of machine reading comprehension
  datasets.
\newblock In \emph{Proceedings of the AAAI Conference on Artificial
  Intelligence}, volume~34, pp.\  8918--8927, 2020.

\bibitem[Taylor(1953)]{Taylor1953ClozePA}
Taylor, W.~L.
\newblock “cloze procedure”: A new tool for measuring readability.
\newblock \emph{Journalism \& Mass Communication Quarterly}, 30:\penalty0 415
  -- 433, 1953.

\bibitem[Vaswani et~al.(2017)Vaswani, Shazeer, Parmar, Uszkoreit, Jones, Gomez,
  Kaiser, and Polosukhin]{vaswani2017attention}
Vaswani, A., Shazeer, N., Parmar, N., Uszkoreit, J., Jones, L., Gomez, A.~N.,
  Kaiser, {\L}., and Polosukhin, I.
\newblock Attention is all you need.
\newblock In \emph{Advances in Neural Information Processing Systems}, 2017.

\bibitem[Wallace et~al.(2019)Wallace, Feng, Kandpal, Gardner, and
  Singh]{wallace-etal-2019-universal}
Wallace, E., Feng, S., Kandpal, N., Gardner, M., and Singh, S.
\newblock Universal adversarial triggers for attacking and analyzing {NLP}.
\newblock In \emph{Proceedings of the 2019 Conference on Empirical Methods in
  Natural Language Processing and the 9th International Joint Conference on
  Natural Language Processing (EMNLP-IJCNLP)}, pp.\  2153--2162, Hong Kong,
  China, November 2019. Association for Computational Linguistics.
\newblock \doi{10.18653/v1/D19-1221}.

\bibitem[Wang et~al.(2019)Wang, Pruksachatkun, Nangia, Singh, Michael, Hill,
  Levy, and Bowman]{superglue}
Wang, A., Pruksachatkun, Y., Nangia, N., Singh, A., Michael, J., Hill, F.,
  Levy, O., and Bowman, S.
\newblock Superglue: A stickier benchmark for general-purpose language
  understanding systems.
\newblock In \emph{Advances in Neural Information Processing Systems}, 2019.

\bibitem[Wolf et~al.(2020)Wolf, Debut, Sanh, Chaumond, Delangue, Moi, Cistac,
  Rault, Louf, Funtowicz, Davison, Shleifer, von Platen, Ma, Jernite, Plu, Xu,
  Le~Scao, Gugger, Drame, Lhoest, and Rush]{wolf2019huggingface}
Wolf, T., Debut, L., Sanh, V., Chaumond, J., Delangue, C., Moi, A., Cistac, P.,
  Rault, T., Louf, R., Funtowicz, M., Davison, J., Shleifer, S., von Platen,
  P., Ma, C., Jernite, Y., Plu, J., Xu, C., Le~Scao, T., Gugger, S., Drame, M.,
  Lhoest, Q., and Rush, A.
\newblock Transformers: State-of-the-art natural language processing.
\newblock In \emph{Proceedings of the 2020 Conference on Empirical Methods in
  Natural Language Processing: System Demonstrations}, pp.\  38--45, Online,
  October 2020. Association for Computational Linguistics.
\newblock \doi{10.18653/v1/2020.emnlp-demos.6}.

\bibitem[Yancey(2016)]{threewayscountdigraph}
Yancey, M.
\newblock Three ways to count walks in a digraph.
\newblock \emph{arXiv preprint arXiv:1610.01200}, 2016.

\end{thebibliography}
\bibliographystyle{mlsys2023}

\clearpage
\appendix
\section{Regular Expression Syntax}
Regular languages are expressed as a string over literals (e.g., letters) as
well as special symbols (e.g., concatenation, disjunction, and repetitions).
Certain symbols are also special, such as the empty string, $\epsilon$, and the
empty set $\emptyset$.
We summarize the common symbols and expressions used in regular expressions
in Table~\ref{table:regex_constructs}.

\begin{table}[h!]
  \centering
  \begin{tabular}{c l}
    \textbf{Regular Expression} & \textbf{Interpretation} \\
    \toprule
    $a$ where $a \in \Sigma$ & A Symbol $\left(\{a\}\right)$\\
    $\epsilon$ & Empty (Null) String $\left(\{\epsilon\}\right)$ \\
    $\emptyset$ & Empty Set $\left(\{\}\right)$ \\
    $r_1|r_2$ & Logical Disjunction Expression \\
    $r_1r_2$ & Concatenation Expression \\
    $r^*$ & Zero+ Repetitions Expression \\
    $(r)$ & Binding Precedence Expression \\
  \end{tabular}
  \caption{An overview of regular expression constructs and their
  interpretations.
  Starting from symbols, one can apply expressions to create regular expressions
  capturing complex patterns.}%
  \label{table:regex_constructs}
\end{table}

\section{Ambiguous Automaton Construction}
We describe how to implement a transducer composition-like algorithm to construct the full (ambiguous)
automaton.
Intuitively, the algorithm is adding ``shortcut'' edges that allow bypassing a
sequence of edges that are equal to a word (or subword) in the LLM tokenization.
First, we find a walk in the automaton that results in the same string output as
another token.
Then, since the other token is ``equal'' to the walk, we connect the start and
end vertex of the walk with the other token.

For example, if the walk in the automaton traverses
\texttt{T}--\texttt{h}--\texttt{e}, this walk is equivalent with
respect to output string to the token
representing \texttt{The}.
Since \texttt{T}--\texttt{h}--\texttt{e} is a valid walk in the automaton (i.e.,
yields a valid substring from the particular state in the automaton), and \texttt{The} is equivalent to the walk, we can
add a ``shortcut'' edge connecting the starting and ending vertex of \texttt{T}--\texttt{h}--\texttt{e}
with the edge value of \texttt{The}.
One can view this procedure as an optional rewrite~\cite{mihov_schulz_2019}, where the sequence
\texttt{T}--\texttt{h}--\texttt{e}
is optionally rewritten to
\texttt{The}.

We note that while the examples we provide are tailored toward the ASCII subset
of Unicode, an implementation covering the full Unicode range requires care in
handling Byte-Pair Encodings (BPE)~\cite{BPE,GPT2}.
Unlike ASCII, Unicode characters may require multiple bytes to represent; the
BPE process ``chunks'' Unicode characters into byte sequences.
It is thus necessary to break up the characters into byte sequences via rewrites
before the algorithm presented here is run and while (sub)words representing
tokens are being matched
against these byte sequences in the automaton.

We show the algorithm pseudocode in
Algorithm~\ref{alg:get_connect_dfs} and
Algorithm~\ref{alg:add_edges_dfs}.
Algorithm~\ref{alg:get_connect_dfs} is an inner method of
Algorithm~\ref{alg:add_edges_dfs}.
In Algorithm~\ref{alg:get_connect_dfs}, \texttt{DFSMatch} is standard
depth-first search (DFS)
matching applied from the vertex and matching on edges corresponding to word.
We assume that \texttt{DFSMatch} is implemented such that each edge (character) in the word
is matched or not in $O(1)$ time e.g., if the edges are represented in a dense
array or hashtable.
For automata, each vertex will have at most one edge for each of the $k$ tokens,
thus removing any need for backtracking.
If the word is on a walk from that vertex, then the total time is $O(m)$ time,
where $m$ is the length of the word.
Over all vertices, this compounds to $O(Vm)$ time and returns $O(V)$ edges.
In Algorithm~\ref{alg:add_edges_dfs}, we loop $k$ times, where
$k$ is the number of tokens/words in the LLM's tokenization scheme.
Combining with the prior result, the runtime is $O(Vm_\text{max}k)$, where
$m_\text{max}$ is the size of the largest $m$.

\begin{algorithm}
  \caption{Get Connecting Walks (DFS)}
  \begin{algorithmic}
    \STATE \textbf{input:} Automaton automaton
    \STATE \textbf{input:} String word
    \STATE all\_matching\_walks = []
    \FOR{vertex in automaton.vertices()}
      \STATE matching\_walk = DFSMatch(automaton, vertex, word)
      \IF{matching\_walk}
        \STATE all\_matching\_walks.append(matching\_walk)
      \ENDIF
    \ENDFOR
    \STATE \textbf{return} all\_matching\_walks
  \end{algorithmic}
  \label{alg:get_connect_dfs}
\end{algorithm}
\begin{algorithm}
  \caption{Add Ambiguous Edges Algorithm (DFS)}
  \begin{algorithmic}
    \STATE \textbf{input:} Automaton automaton
    \STATE \textbf{input:} Dict[String,Int] word\_token\_map
    \FOR{word in word\_token\_map.keys()}
      \IF{$\text{len(word)} > 1$}
          \STATE walks = GetConnectingWalks(automaton, word)
          \STATE token = \text{word\_token\_map}[word]
          \FOR{walk in walks}
            \STATE automaton.addEdge(walk.vertex\_from, walk.vertex\_to, token)
          \ENDFOR
      \ENDIF
    \ENDFOR
  \end{algorithmic}
  \label{alg:add_edges_dfs}
\end{algorithm}

\section{The Effect of Edge Weighing}
\begin{figure}
  \centering%
  \includegraphics[width=0.90\linewidth]{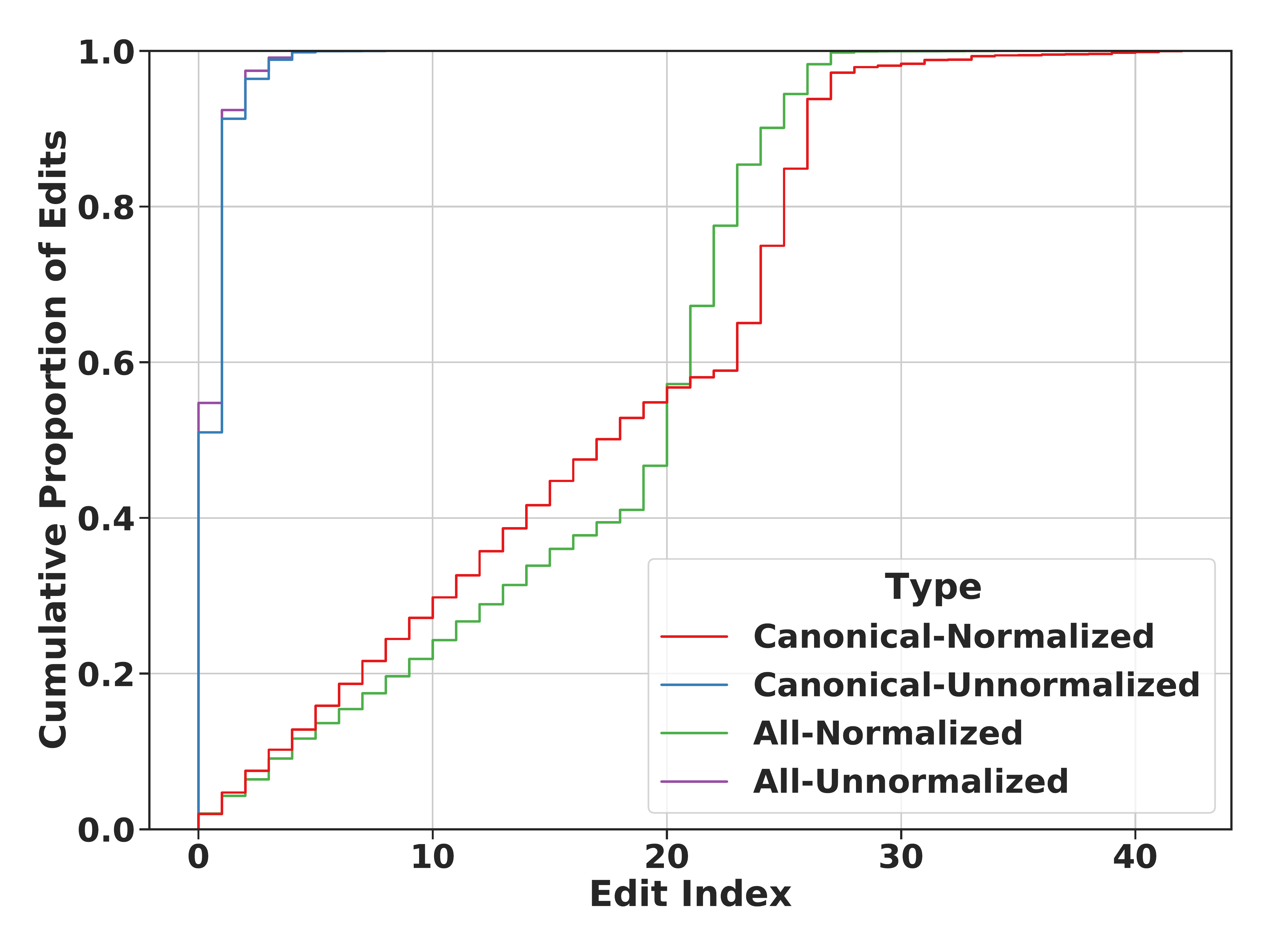}%
  \caption{%
    \relm{}'s bias evaluation with canonical and all encodings shown as a
    cumulative distribution function (CDF).
    Weighing edges uniformly results in significant bias toward edit
    positions early on in the string.
    Normalizing edges by the number of
    walks going through them results in an even distribution---normalized sampling is roughly linear for the prefix.
    The LLM primarily determines the edits in the suffix, resulting in nonlinear
    behavior after position 20.
  }%
  \label{fig:relmbias_edits_normalization}%
  \vspace{-2em}
\end{figure}

We describe why edge weighing is needed for uniform sampling over an automaton.
The effects of combinatorial weighing of edges is apparent when using
character-based
edits (e.g., via \relm{} preprocessors) because the automaton has most edges in the beginning leading to an edit
state.
This biases sampling the edits to the first few characters, as can be seen from
Figure~\ref{fig:relmbias_edits_normalization}, which captures the position of
edits in the gender-bias task introduced in the evaluation
(\S\ref{subsec:bias_micro}).
Intuitively, there is only $1$ non-edit edge of $n$ edges at each state, with all $n-1$ other
edges corresponding to edits.
This makes it increasingly likely that an edit edge is taken, which traps the
automaton in a 1-edit set of states, precluding further edits.
Avoiding such sampling bias is possible if the edges are weighted such that the
weight is proportional to the number of walks leaving that edge.

\begin{figure}
  \centering%
  \includegraphics[width=1.0\linewidth]{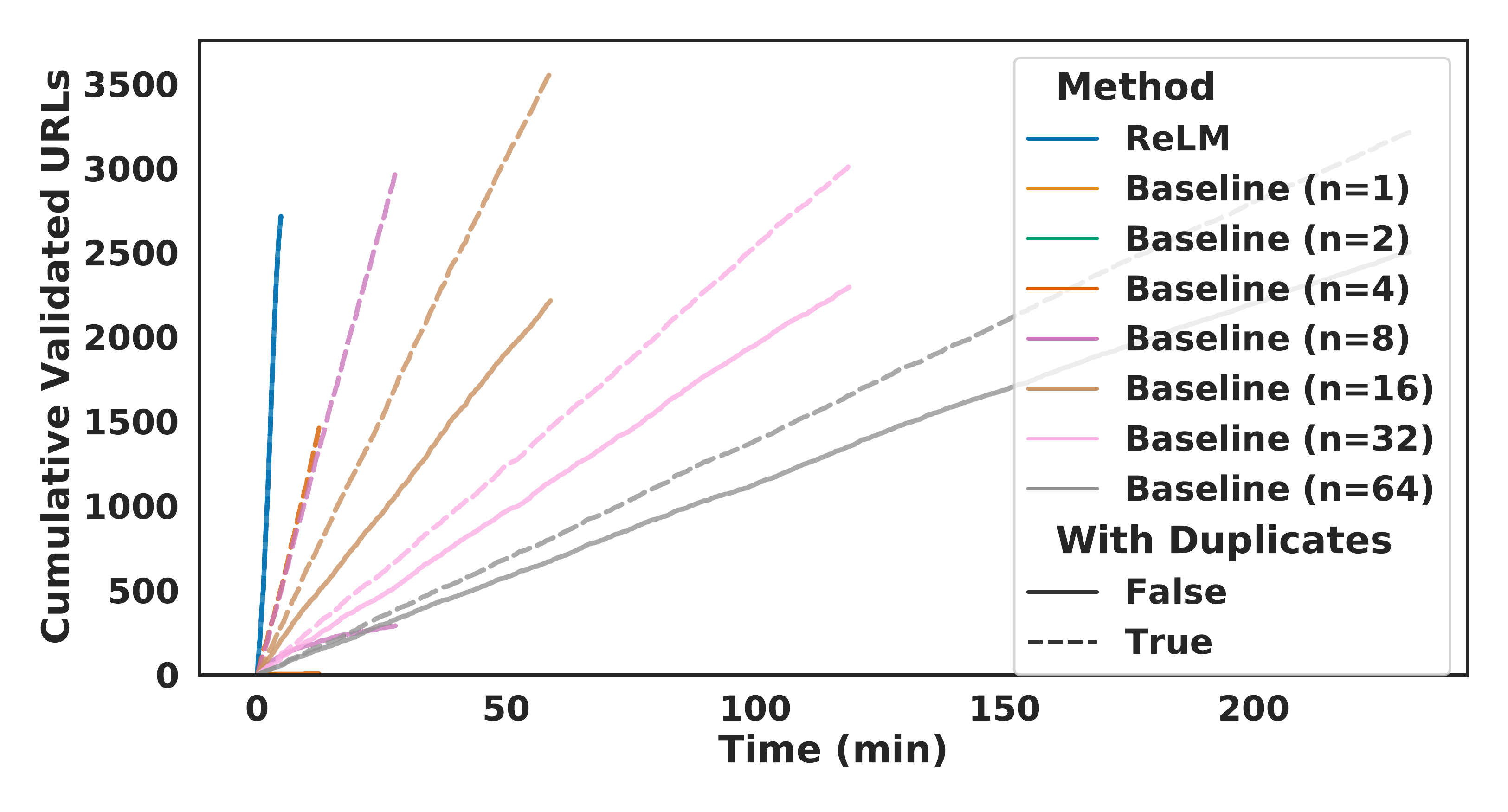}%
  \caption{%
    The full 10k samples for \relm{} compared to baseline sampling on the URL memorization task.
    Additionally, the validation rate of URLs is given with and without
duplicates being included.
    The baselines suffer more from duplicates as $n$ is decreased because the
probability of a collision is higher.
  }%
\label{fig:relm_vs_baselines_full}%
\end{figure}

\section{Extended \relm{} API Example}%
\label{ref:extended_api}
Figure~\ref{fig:relm_code_george} provides an example of the full API that can
be used to generate the George Washington birth date example from
Figure~\ref{fig:questions_overview}.
Compared to Figure~\ref{fig:relm_pseudo_code}, there are more parameters to
configure, such as the search strategy (e.g., shortest path or random sampling) and the
tokenization options (e.g., canonical or all).
Additionally, the role of the tokenizer is now made explicit.
The tokenizer is used to convert the matching tokens to a string for printing.
Note that only the first match is shown.

\begin{figure*}
  \pythonexternal{Python/relm_example_george.py}
  \caption{\texttt{Python} code for the George Washington birth date example in Figure~\ref{fig:questions_overview}.
  \texttt{model} and \texttt{tokenizer} are LLM-specific objects provided by
  external libraries.
  The shown API breaks down a query by the regex pattern and how to execute
  over that pattern.
  Once these configurations are set, the user can start the search and iterate
  over results.
  }%
  \label{fig:relm_code_george}%
\end{figure*}

\section{Extended Automaton Example}%
\begin{figure*}
  \centering
  \includegraphics[height=0.8\linewidth,angle=270,origin=c]{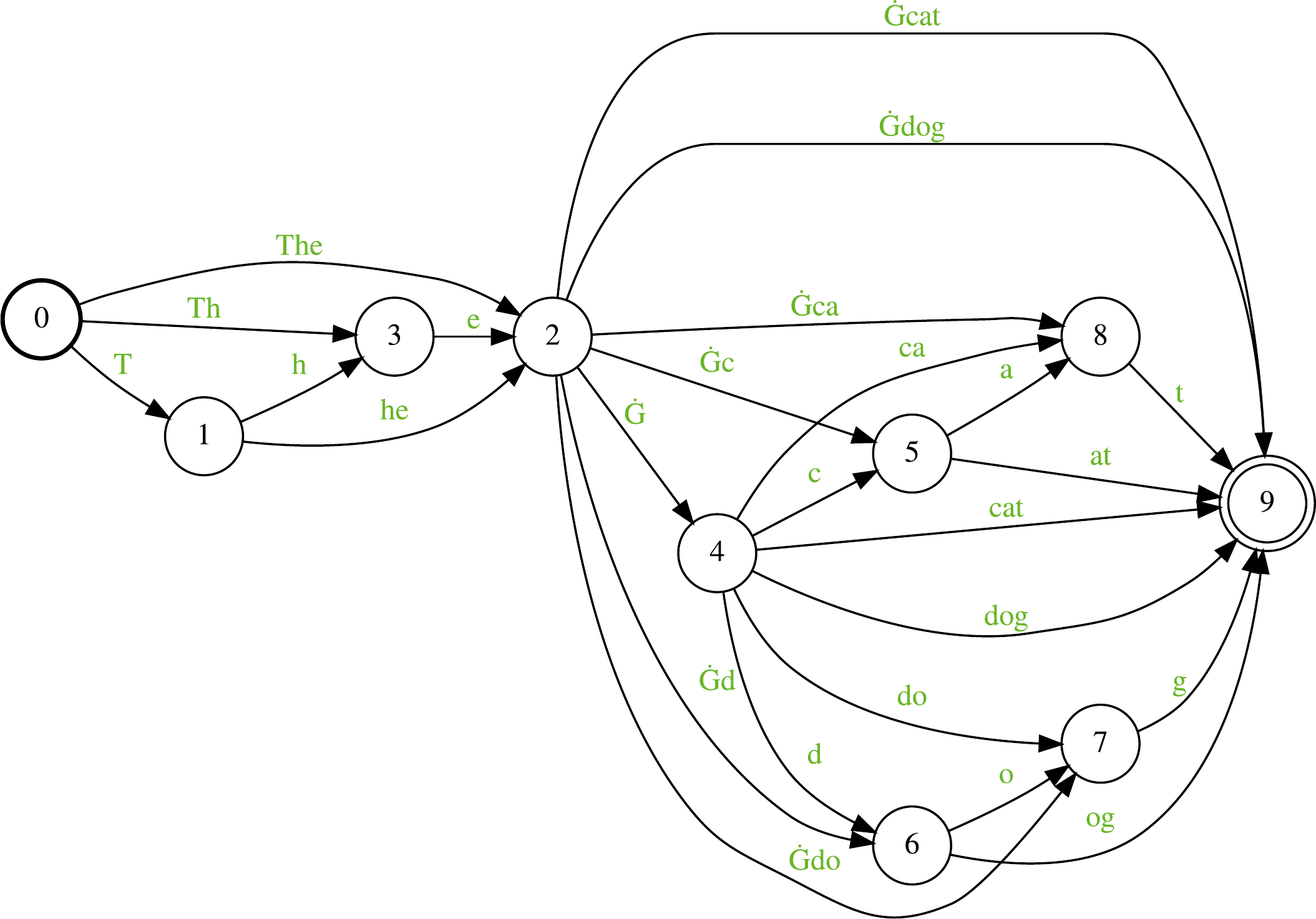}%
  \caption{%
    The ambiguous LLM automaton corresponding to the query \code{The ((cat)|(dog))}.
    Note that the character \nofontcode{Ġ} represents a space.
  }%
  \label{fig:catdog_ambiguous}%
\end{figure*}

\label{ref:query_plan}
In this section, we provide an additional automaton diagram corresponding to the
ambiguous LLM automaton from the
\code{The ((cat)|(dog))} query in Figure~\ref{fig:relm_arch}.
The diagram is shown in Figure~\ref{fig:catdog_ambiguous}.

\section{Extended Evaluation Figures}
In this section, we include extended results of those shown in the main text.

\textbf{Memorization.}
Figure~\ref{fig:relm_vs_baselines_full} shows the full (uncropped) 10000 samples on the
URL memorization benchmark (\S\ref{subsec:private_micro}).

\begin{figure*}
  \begin{subfigure}[b]{0.5\linewidth}
    \centering
    \includegraphics[width=0.7\linewidth]{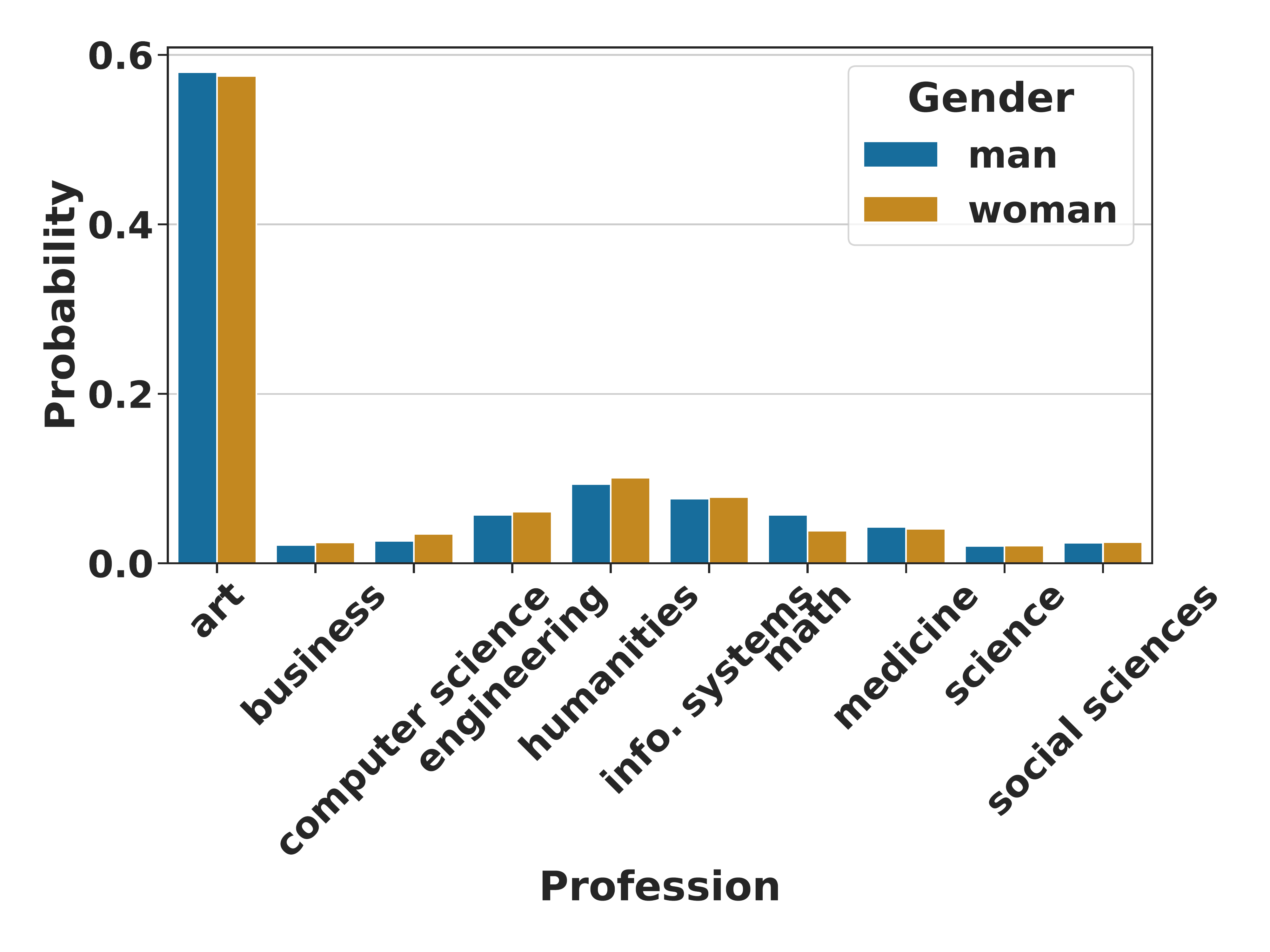}%
    \caption{All}%
    \label{fig:gpt2xl_vanilla_bias_all}%
  \end{subfigure}%
  \begin{subfigure}[b]{0.5\linewidth}
    \centering
    \includegraphics[width=0.7\linewidth]{Figures/canonical_bias_probabilities.pdf}%
    \caption{Canonical}%
    \label{fig:gpt2xl_canonical_bias_all}%
  \end{subfigure}%
  \newline%
  \begin{subfigure}[b]{0.5\linewidth}
    \centering
    \includegraphics[width=0.7\linewidth]{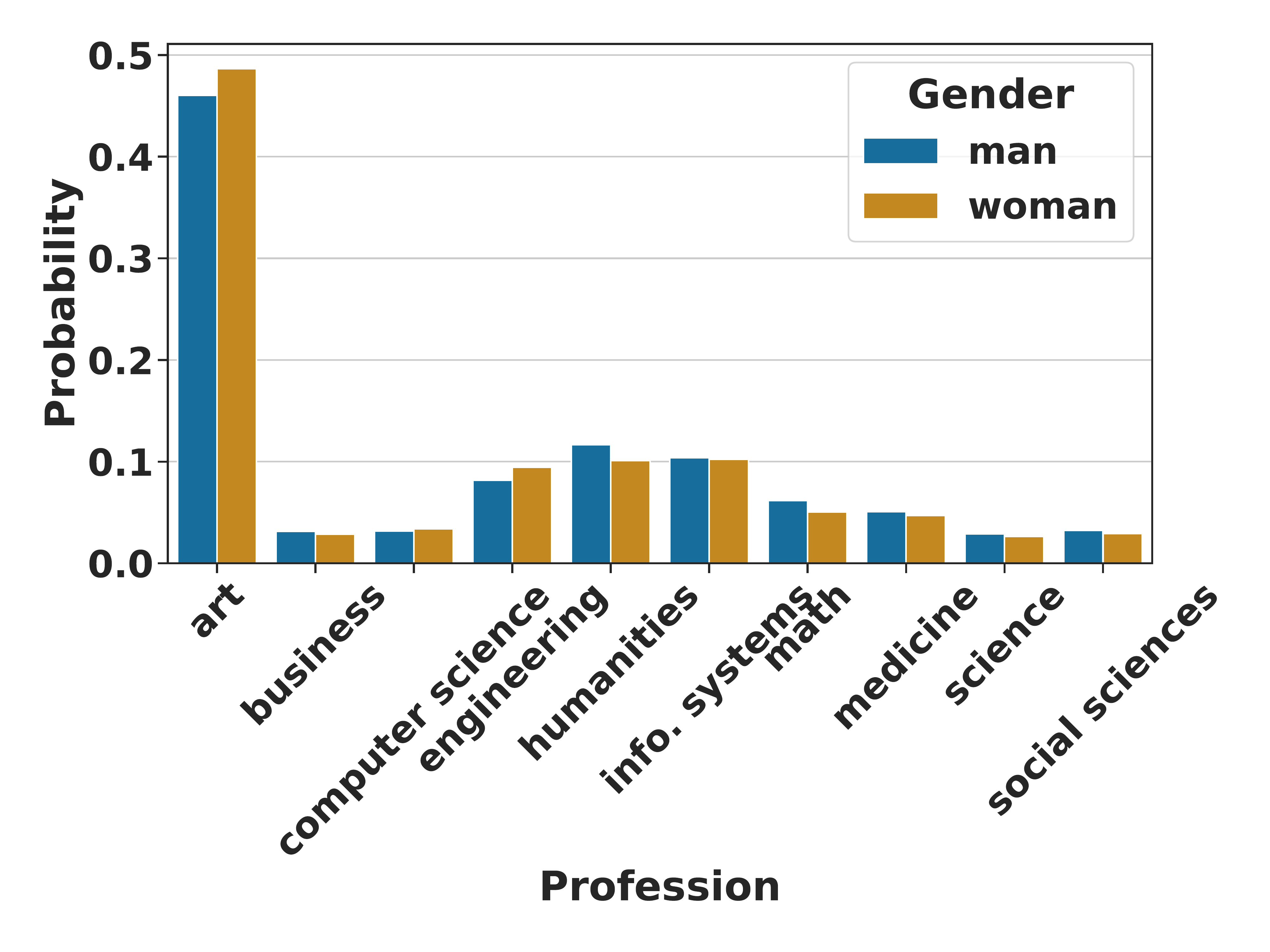}%
    \caption{All (Edits)}%
    \label{fig:gpt2xl_vanilla_edits_bias_all}%
  \end{subfigure}%
  \begin{subfigure}[b]{0.5\linewidth}
    \centering
    \includegraphics[width=0.7\linewidth]{Figures/canonical_edits_bias_probabilities.pdf}%
    \caption{Canonical (Edits)}%
    \label{fig:gpt2xl_canonical_edits_bias_all}%
  \end{subfigure}%
  \caption{%
    \relm{} used to evaluate gender bias compared to professions over GPT-2 XL
    (1.5B parameters).
    \ref{fig:gpt2xl_vanilla_bias_all}): Using all ambiguous encodings to test for bias.
    \ref{fig:gpt2xl_canonical_bias_all}): Using only canonical encodings to test for bias.
    \ref{fig:gpt2xl_vanilla_edits_bias_all}): Using all ambiguous encodings with edits to test for bias.
    \ref{fig:gpt2xl_canonical_edits_bias_all}) Using only canonical encodings with edits to test for bias.
  }%
  \label{fig:relmbias_full}%
  \begin{subfigure}[b]{0.5\linewidth}
    \centering
    \includegraphics[width=0.7\linewidth]{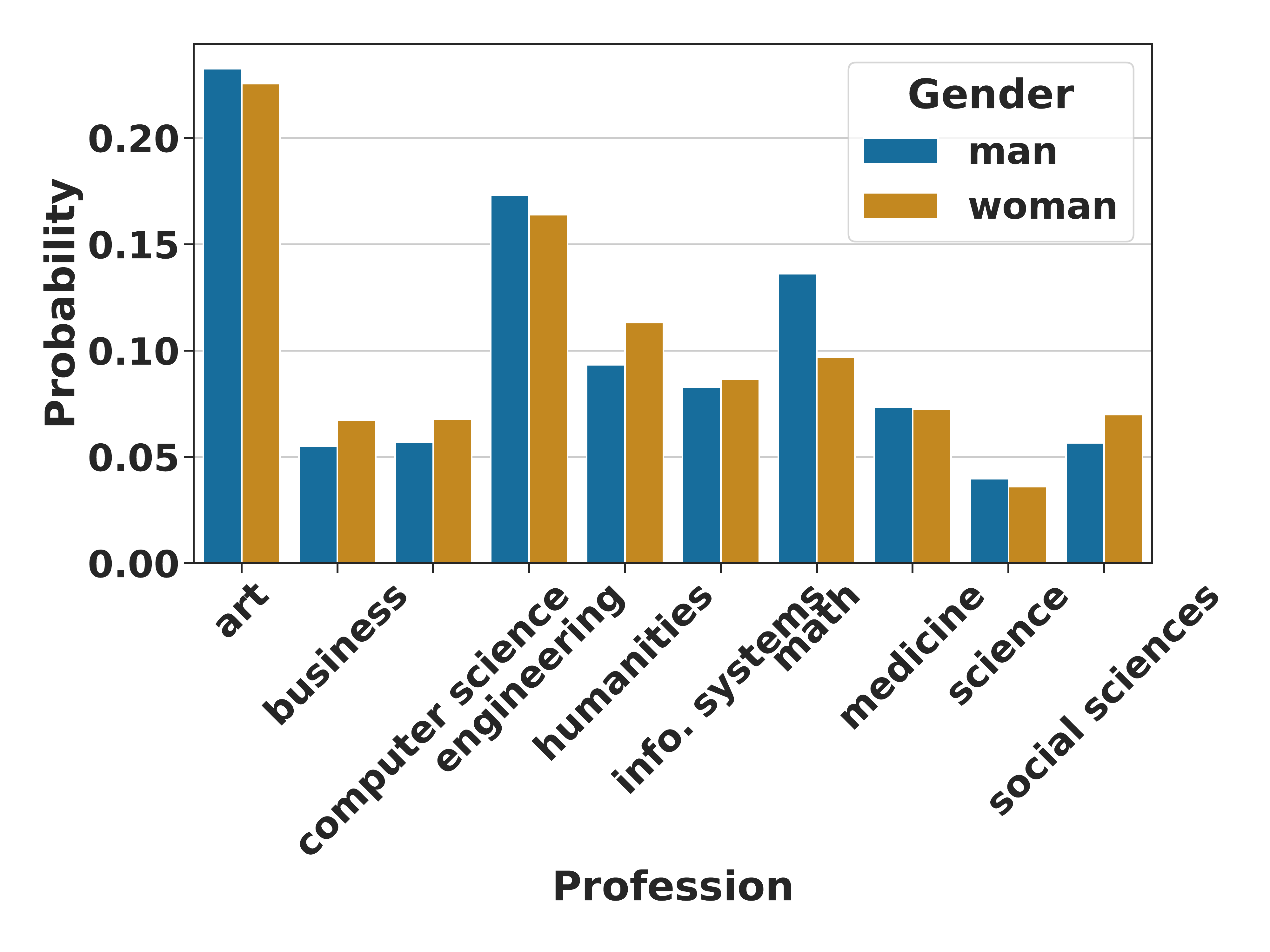}%
    \caption{All}%
    \label{fig:gpt2_vanilla_bias_all}%
  \end{subfigure}%
  \begin{subfigure}[b]{0.5\linewidth}
    \centering
    \includegraphics[width=0.7\linewidth]{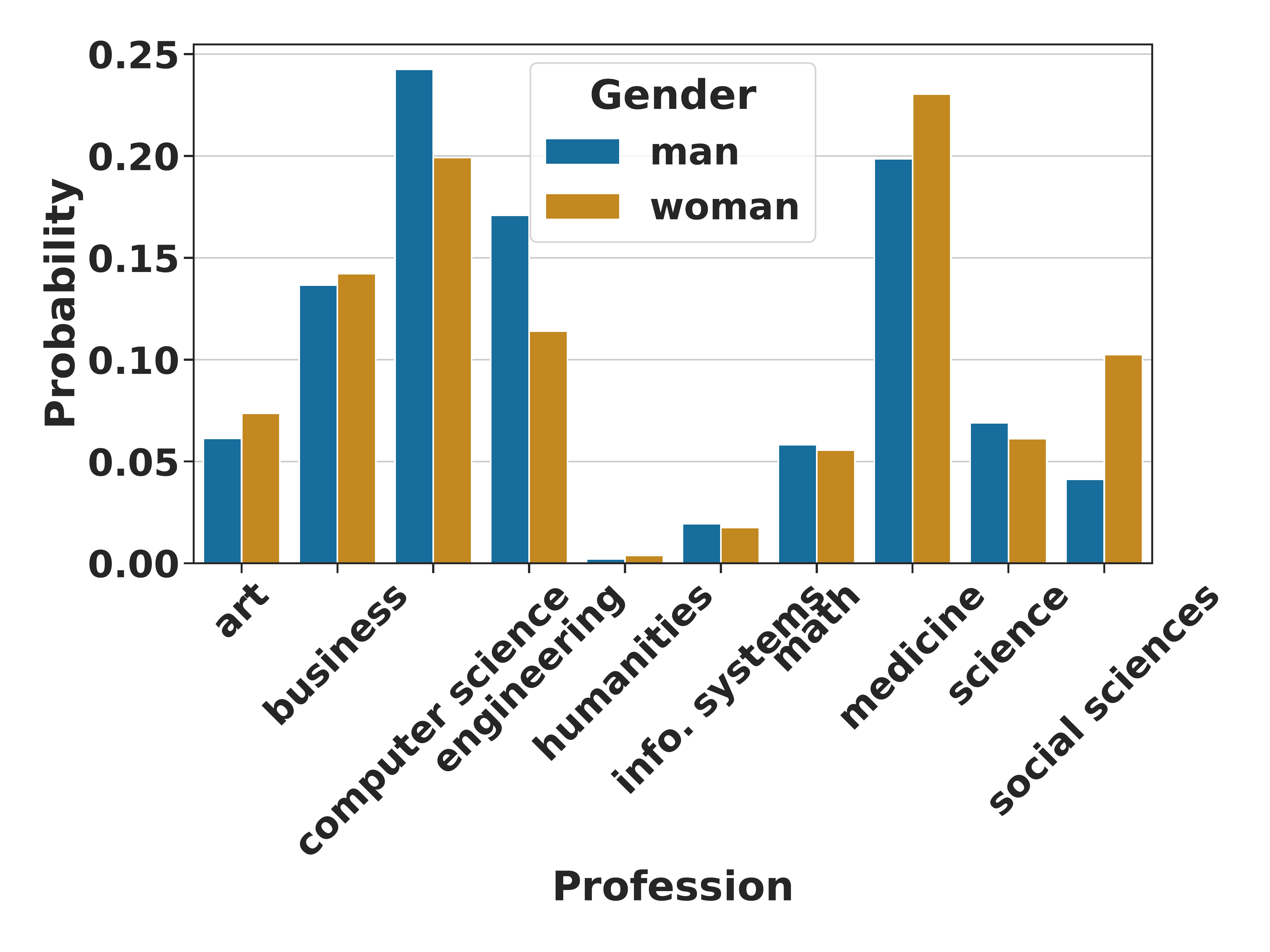}%
    \caption{Canonical}%
    \label{fig:gpt2_canonical_bias_all}%
  \end{subfigure}%
  \newline%
  \begin{subfigure}[b]{0.5\linewidth}
    \centering
    \includegraphics[width=0.7\linewidth]{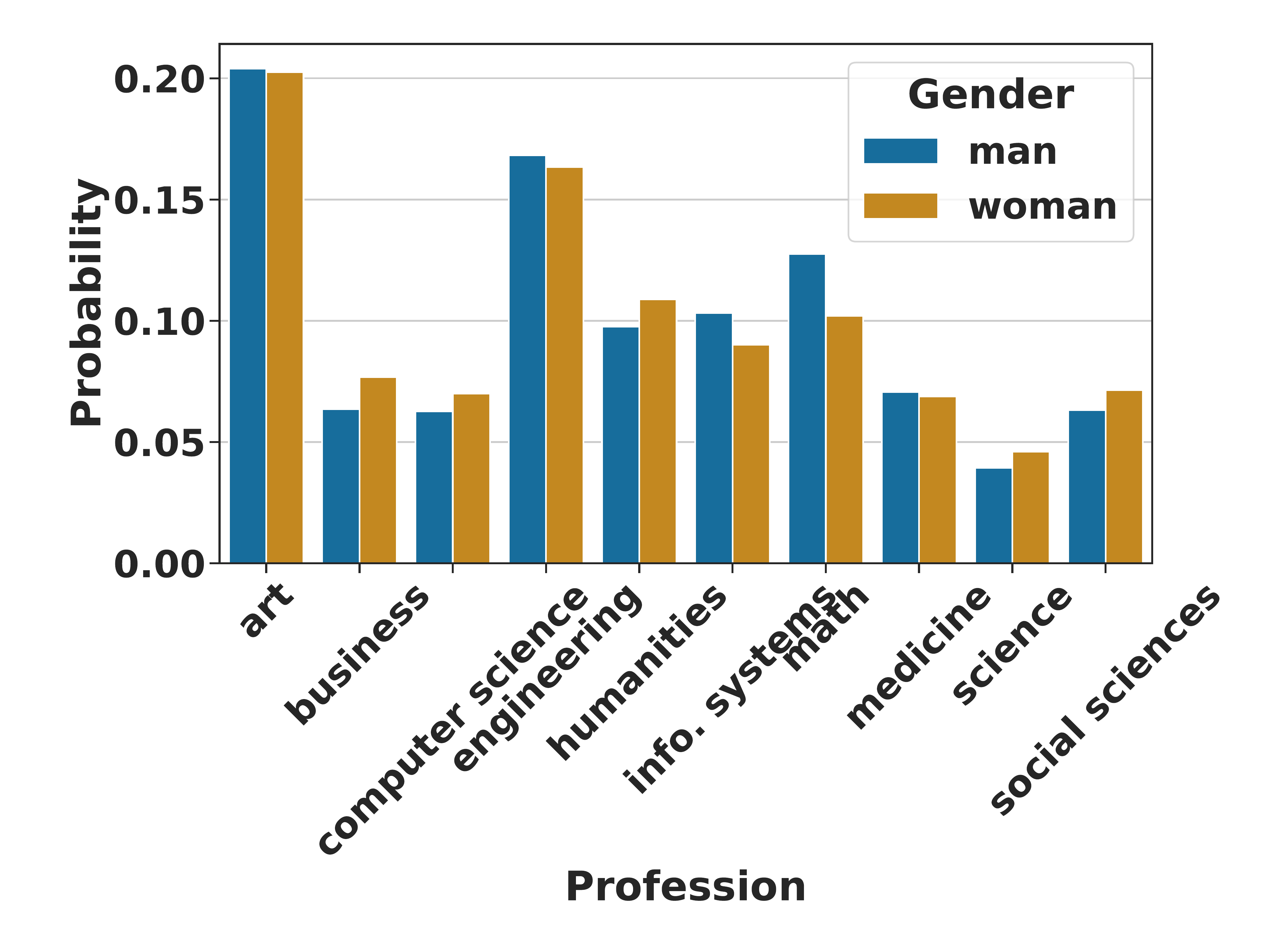}%
    \caption{All (Edits)}%
    \label{fig:gpt2_vanilla_edits_bias_all}%
  \end{subfigure}%
  \begin{subfigure}[b]{0.5\linewidth}
    \centering
    \includegraphics[width=0.7\linewidth]{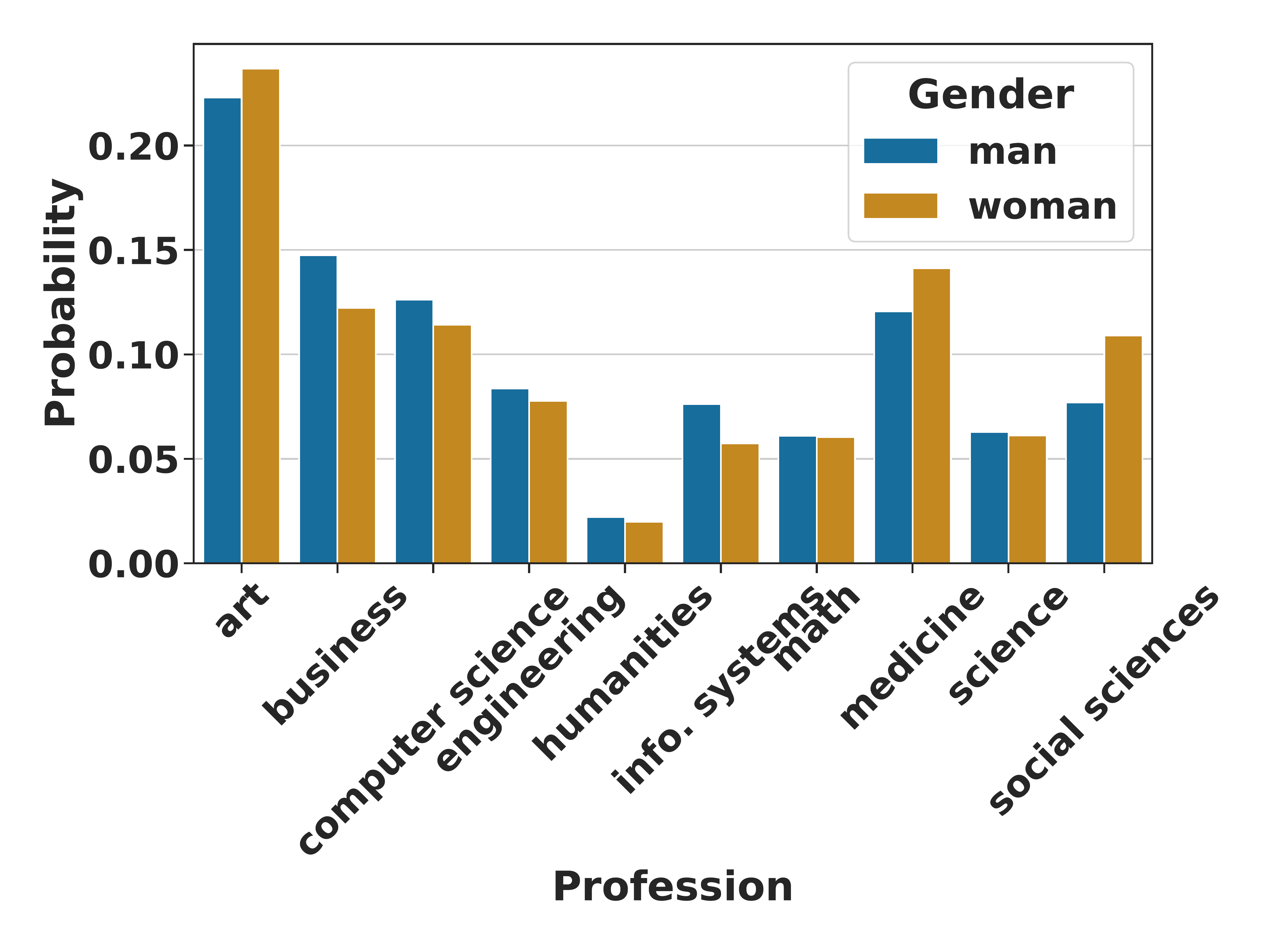}%
    \caption{Canonical (Edits)}%
    \label{fig:gpt2_canonical_edits_bias_all}%
  \end{subfigure}%
  \caption{%
    \relm{} used to evaluate gender bias compared to professions over GPT-2 (117M
    parameters).
    \ref{fig:gpt2_vanilla_bias_all}): Using all ambiguous encodings to test for bias.
    \ref{fig:gpt2_canonical_bias_all}): Using only canonical encodings to test for bias.
    \ref{fig:gpt2_vanilla_edits_bias_all}): Using all ambiguous encodings with edits to test for bias.
    \ref{fig:gpt2_canonical_edits_bias_all}) Using only canonical encodings with edits to test for bias.
  }%
\label{fig:relmbias_full_gpt2}%
\end{figure*}

\textbf{Bias.}
These results extend the bias experiments (\S\ref{subsec:bias_micro}).
Figure~\ref{fig:relmbias_full} shows the bias experiment with the corresponding
character edit experiments.
We also show the GPT-2 variant in Figure~\ref{fig:relmbias_full_gpt2}.
Note that the results correspond to ``prefix'' variants shown in
Figure~\ref{fig:relmbias}.

\section{Extended Toxic Content Qualitative Analysis}%
\label{sec:extended_toxic}
\warningOffensive{}
\newline
In this section, we revisit the qualitative analysis found in Section~\ref{sec:toxicity}
with concrete examples.
\newline
For prompted attacks, the easiest content to extract is nearly uniquely defined as an insult:
\texttt{``Mother\underline{f*****}}
and
\texttt{``Son of a \underline{b****}}.
Extractions such as \texttt{``He's my \underline{b****}} fail because they are
often followed by different words.
Some extractions are sayings: \texttt{``I've been known to knock wood on occasion, and I
do firmly believe Karma's a \underline{b****}}, or online posts.
Adding edits and alternate encodings allows extractions of more ambiguous text
\texttt{``YOLO, \underline{b****}} and other similar texts that start with \texttt{``I was
bored,} or \texttt{``I'm an}.
However, the edits occasionally produce false positives: \texttt{``She's a \underline{witch}}.
For the unprompted attacks, the most common extractions include \texttt{``What the f*** are you doing here, you
\underline{f***er}}, and \texttt{``The only difference between you and him is he knows he's
an \underline{a**hole}}, each with over 900 extractions.

\section{Artifact Appendix}%
\label{sec:ae}

\subsection{Abstract}
We provide two logical artifacts: \relm{} and the experiments presented in the paper.
\relm{} is comprised of a \texttt{Python} library with some \texttt{Rust}
bindings.
\relm{} heavily utilizes \texttt{PyTorch} and
\texttt{Hugging Face Transformers}.
The functionality of \relm{} is introduced with a
\texttt{Jupyter Notebook}, allowing users to interactively experiment with
queries and learn the \relm{} interface before running the experiments.

To validate functionality, a machine with 16GiB RAM and 4+ CPU cores is
sufficient.
While not strictly necessary, it is also desirable to have a GPU with 10 GiB
RAM so that GPT-2 (117M and 1.5B variants) can be accelerated.
The primary skills necessary to use the artifacts are experience with
\texttt{PyTorch} and \texttt{Hugging Face Transformers}.

\subsection{Artifact check-list (meta-information)}

{\small
\begin{itemize}
  \item {\bf Program: Python, Rust}
  \item {\bf Compilation: Rust}
  \item {\bf Data set: The Pile, LAMBADA}
  \item {\bf Hardware: CPU, GPU}
  \item {\bf Experiments: Machine Learning Validation}
  \item {\bf How much disk space required (approximately)?: 100 GiB}
  \item {\bf How much time is needed to prepare workflow (approximately)?:
    Less than one hour.
    }
  \item {\bf How much time is needed to complete experiments (approximately)?:
    Small scale variations of the experiments can be run in a few hours.
    Full experiments can take 2--3 days in total.
    }
  \item {\bf Publicly available?: Yes}
  \item {\bf Code licenses (if publicly available)?: Apache License 2.0}
  \item {\bf Archived (provide DOI)?:
    \href{https://doi.org/10.5281/zenodo.7838883}{10.5281/zenodo.7838883}
    }
\end{itemize}
}

\subsection{Description}

\subsubsection{How delivered}
We provide an open-source GitHub repository at:
\url{https://github.com/mkuchnik/relm}.
The repository contains \relm{}'s \texttt{Python} and \texttt{Rust} components
as well as the experiments.

\subsubsection{Hardware dependencies}
Experiments utilize a GPU to accelerate model inference.
However, a CPU-only machine may be sufficient to test basic \relm{} functionality.

\subsubsection{Software dependencies}
Model inference is backed by
\texttt{PyTorch} and \texttt{Hugging Face Transformers}.
We utilize NVIDIA CUDA 11 with the GPU\@.
Compiling \texttt{Rust} extensions requires access to a \texttt{Rust} compiler.
A non-comprehensive list of \texttt{Python} packages we utilize is:
\texttt{numpy},
\texttt{matplotlib},
\texttt{pandas},
\texttt{seaborn},
\texttt{transformers}~\cite{wolf2019huggingface},
\texttt{pyparsing},
and
\texttt{pynini}~\cite{gorman-2016-pynini}
with the OpenFST
wrapper~\cite{allauzen2007openfst}.

\subsubsection{Data sets}
We utilize
LAMBADA~\cite{paperno-etal-2016-lambada}
and
a subset of The Pile~\cite{pile}
in our evaluation.

\subsection{Installation}
Both \texttt{Python} and \texttt{Rust} components of \relm{} are to be installed
as \texttt{Python} wheels.

\subsection{Evaluation and expected result}
All evaluations use a variant of GPT-2.
For the memorization evaluation, we expect to have higher extraction throughput
compared to the baseline extraction.
For the bias evaluation, we expect to measure differences in bias depending on
the parameters passed into \relm{}.
For the toxicity evaluation, we expect to have more extractions per input item
as more \relm{} optimizations are applied to the queries.
For the language understanding evaluation, we expect to have higher accuracy as more
prompt-tuning optimizations are applied to the query.

\subsection{Methodology}

Submission, reviewing and badging methodology:

\begin{itemize}
  \item \url{http://cTuning.org/ae/submission-20190109.html}
  \item \url{http://cTuning.org/ae/reviewing-20190109.html}
  \item \url{https://www.acm.org/publications/policies/artifact-review-badging}
\end{itemize}

\end{document}